\documentclass{article}
\PassOptionsToPackage{numbers}{natbib}
 
\usepackage[margin=1.5in]{geometry} 
\usepackage{times} 

\usepackage{authblk} 
\usepackage{natbib}

\usepackage{amsmath,amsfonts}
\usepackage{graphicx}
\usepackage{textcomp}
\usepackage{xcolor}
\usepackage{multirow}
\usepackage{url}
\PassOptionsToPackage{hyphens}{url}
\usepackage{hyperref}
\usepackage{makecell}
\usepackage{multirow}
\usepackage{graphicx}
\usepackage{wrapfig}
\graphicspath{ {images/} }
\usepackage{color}
\usepackage{booktabs}

\usepackage{algorithm}
\usepackage[noend]{algpseudocode}

\usepackage{adjustbox}

\algnewcommand{\LineComment}[1]{\State \(\triangleright\) #1}

\usepackage[caption=false,font=footnotesize]{subfig}

\usepackage{titling}

\title{Boosting Deep Ensembles with Learning Rate Tuning}

\author{%
  Hongpeng Jin \quad Yanzhao Wu \\
  Florida International University \\
  Miami, FL, United States \\
  \texttt{\{hjin008, yawu\}@fiu.edu} \\
}

\date{} 

\begin{document}

\maketitle

\begin{abstract} 
The Learning Rate (LR) has a high impact on deep learning training performance. A common practice is to train a Deep Neural Network (DNN) multiple times with different LR policies to find the optimal LR policy, which has been widely recognized as a daunting and costly task. Moreover, multiple times of DNN training has not been effectively utilized. In practice, often only the optimal LR is adopted, which misses the opportunities to further enhance the overall accuracy of the deep learning system and results in a huge waste of both computing resources and training time. This paper presents a novel framework, LREnsemble, to effectively leverage effective learning rate tuning to boost deep ensemble performance. We make three original contributions.
First, we show that the LR tuning with different LR policies can produce highly diverse DNNs, which can be supplied as base models for deep ensembles.
Second, we leverage different ensemble selection algorithms to identify high-quality deep ensembles from the large pool of base models with significant accuracy improvements over the best single base model.
Third, we propose LREnsemble, a framework that utilizes the synergy of LR tuning and deep ensemble techniques to enhance deep learning performance.
The experiments on multiple benchmark datasets have demonstrated the effectiveness of LREnsemble, generating up to 2.34\% accuracy improvements over well-optimized baselines.

\end{abstract}



\section{Introduction} 



Deep learning has been widely applied in many real-world applications, represented by the recent success of Large Language Models~\cite{arc, bleu, triviaqa, coqa, superglue,gpt3, llama} to achieve human-like performance. 
The Learning Rate (LR) is a critical hyperparameter in deep learning with high impacts on the Deep Neural Network (DNN) training performance~\cite{clr, clr-22, Bengio-PracticalRecommendations}.
Good learning rates can train deep learning models with high accuracy or reach accuracy thresholds fast with low training costs~\cite{LRBenchTIST, adagrad}.
Conversely, too small or too large learning rates may result in slow convergence or model divergence during deep learning training~\cite{clr,Bengio-PracticalRecommendations,LRBenchBigData}.
Hence, finding a good learning rate is challenging, and requires meticulous tuning.
Even though numerous efforts have been devoted to designing learning rate policies to leverage decaying, cyclic, or composite functions to specify LR values during training to facilitate LR tuning~\cite{sgdr,clr,LRBenchTIST}, it still requires multiple times of deep learning training with trial-and-errors to find the optimal LR policy, making the LR tuning not only tedious but also time-consuming. It is still an open research challenge to identify the optimal LR policy.
In practice, often only the optimal LR policy will be selected to train the deep learning model and shared across the deep learning community. For a new learning task, a new dataset, or a new model, the LR tuning process will repeat and cause a huge waste of computing resources. We bring up a novel research question in this study, i.e., how to effectively utilize the sub-optimal LR tuning outcomes to improve deep learning performance.

\begin{wrapfigure}{r}{0.4\textwidth}
  \begin{center}
    \includegraphics[width=0.4\textwidth]{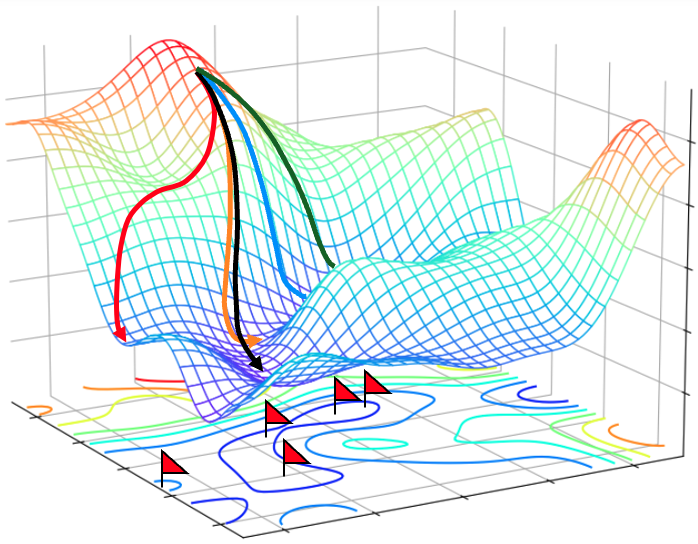}
  \end{center}
  \vspace{-4mm}
  \caption{Visualization of deep learning training paths with different LR policies: different LR policies can lead to different optimization trajectories and produce diverse deep learning models.}
\label{fig:visualization_optimization_path}
\vspace{-1mm}
\end{wrapfigure}

\begin{wrapfigure}{r}{0.56\textwidth} 
\vspace{-8mm}
\centering
\subfloat[\scriptsize LREnsemble vs. Individual models]{
  \centering
  \includegraphics[width=0.275\textwidth]{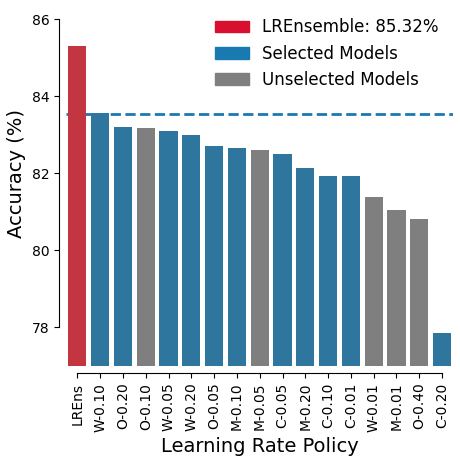}
  \label{fig:first_page_cifar100_lrensemble}
} 
\subfloat[\scriptsize Individual models by LR tuning]{
  \centering
  \includegraphics[width=0.275\textwidth]{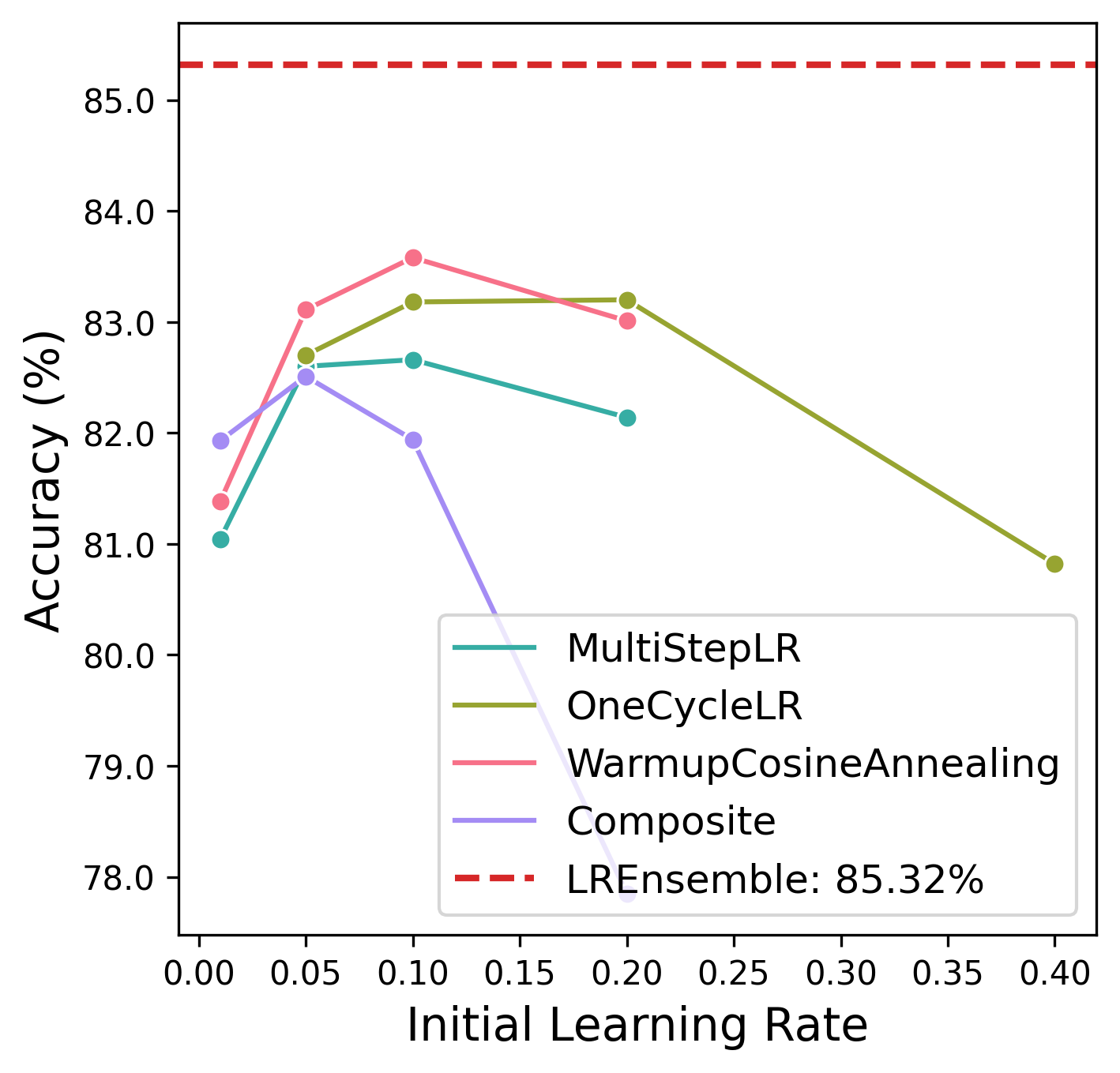}
  \label{fig:first_page_cifar100_lrpolicy}
} 
\vspace{-3mm}
\caption{Accuracy improvements by LREnsemble (WRN-28-10 on CIFAR-100): LREnsemble can leverage Learning Rate (LR) tuning to generate diverse individual models (blue or gray bars in Figure~\ref{fig:first_page_cifar100_lrensemble}) and select the complementary member models (blue bars: selected models) to boost ensemble accuracy (red bar: LREnsemble with over 1.74\% accuracy improvements). Figure~\ref{fig:first_page_cifar100_lrpolicy} shows the accuracy of individual models trained using different LR policies with different initial LR values using our LREnsemble framework.}
\label{fig:first_page_cifar100}
\vspace{-4mm}
\end{wrapfigure}

Deep neural network Ensembles (Deep Ensembles) represent a category of emerging techniques, which hold the potential to combine multiple diverse DNNs to enhance the predictive performance over individual DNNs~\cite{lakshminarayanan2017simple,deepensembles,wu2021boostingdeep,wenzel2020hyperparameter,wu2023hierarchical,wen2020batchensemble}.  
Both good individual model accuracy and high ensemble diversity contribute to the potential accuracy improvements by deep ensembles~\cite{lakshminarayanan2017simple,wen2020batchensemble,wenzel2020hyperparameter,wu2021boostingdeep,diversity-and-generalization-in-nn-ensembles,efficient-diversity-ensemble,wu2023hierarchical}.
LR tuning will employ different LR policies to train DNNs, resulting in different optimization trajectories, each represented by different colors in Figure~\ref{fig:visualization_optimization_path}.
Multiple diverse DNNs produced through this process can be potentially combined through deep ensembles to enhance the overall deep learning accuracy. 
We show the accuracy of individual DNNs obtained with LR tuning in blue or gray bars in Figure~\ref{fig:first_page_cifar100_lrensemble} for training WRN-28-10~\cite{zagoruyko2016wide} on CIFAR-100~\cite{cifar10-100}, where each bar corresponds to a data point in Figure~\ref{fig:first_page_cifar100_lrpolicy} with different LR policies and different initial LR values. We observe that these DNNs have varying model accuracy, exhibiting a certain degree of diversity, making them potential candidates for creating high-quality deep ensembles. In Figure~\ref{fig:first_page_cifar100_lrensemble}, the high-quality ensemble (red bar) achieves the accuracy of 85.32\%, which significantly outperforms the best single DNN of 83.58\% accuracy.
On the other hand, we also found that simply combining all available DNNs will introduce high ensemble execution costs but may not produce optimal accuracy. For example, the red ensemble in Figure~\ref{fig:first_page_cifar100_lrensemble} with 11 member DNNs (blue bars) outperforms the entire ensemble of all 16 DNNs with 85.23\% accuracy. Hence, given the large number of models produced by LR tuning, it is non-trivial to identify high-quality ensembles to improve deep learning accuracy and reduce the ensemble cost.

In this paper, we propose LREnsemble, a novel framework to leverage effective \textbf{LR} tuning to boost deep \textbf{Ensemble} performance, which addresses all three research challenges mentioned above: (1) how to perform effective LR tuning to identify optimal LR policies for training individual DNNs for delivering high accuracy, (2) how to utilize sub-optimal models generated by LR tuning to enhance deep learning accuracy and avoid the waste of computing resources, and (3) how to efficiently build high-quality deep ensembles from these models produced by LR tuning to further enhance overall prediction performance.
We make three novel contributions.
\textit{First,} we show that leveraging different LR policies to perform LR tuning can not only identify optimal LR policies but also produce highly diverse DNNs to facilitate ensemble learning.
\textit{Second,} we demonstrate that the sub-optimal models produced by LR tuning can be effectively leveraged to develop deep ensembles to enhance the overall predictive performance.
\textit{Third,} we introduce LREnsemble, a novel framework that exploits the synergy between LR tuning and deep ensemble to enhance deep learning performance.
We conduct comprehensive experiments on multiple benchmark datasets and DNNs, including ViT~\cite{dosovitskiy2020image} and a representative Large Language Model (LLM), LLaMA~\cite{llama}, to demonstrate the effectiveness of our LREnsemble framework. To the best of our knowledge, this is the first to utilize the sub-optimal models produced by LR tuning to improve deep learning performance.

\section{Related Work}
In this section, we summarize and discuss related studies in three respects: (1) Learning Rate Tuning, (2) Deep Ensembles, and (3) Large Language Models.

\noindent \textbf{Learning Rate Tuning.}
The Learning Rate (LR) is one of the most critical hyperparameters in deep learning training. Different from other static hyperparameters, such as the batch size and the number of DNN layers, LR can be adjusted dynamically during DNN training, which provides high flexibility but also presents challenges in LR tuning. An LR policy ($\eta(t)$) defines the LR value for iteration/epoch $t$. Several existing studies~\cite{LRBenchBigData, LRBenchTIST, jin2023rethinking} present an in-depth summary of representative LR policies and analyze the high impacts of LR tuning on deep learning training performance. Decaying LRs present one category of LR policies defined by a decaying function to start with a high LR value and decrease the LR throughout the training. For example, MultiStepLR is specified by a step function with $l$ indicating multiple step lengths and reducing the LR value by a factor of $\gamma$ each step~\cite{zagoruyko2016wide,wenzel2020hyperparameter}. Another category of popular LR policies is cyclic LRs, which are characterized by periodically increasing and decreasing the LR values during DNN training, represented by CLRs~\cite{clr}, SGDR~\cite{sgdr}, and OneCycleLR~\cite{superconvergence}. OneCycleLR simply uses one cycle from a cyclic function, which will first increase the LR value to its upper bound and then decrease the LR to help the model converge fast close to the end of training. The cyclic LRs have been leveraged in~\cite{huang2017snapshot} to generate multiple DNN snapshots during one training to develop the snapshot ensemble, which is different from the proposed LREnsemble to develop deep ensembles from multiple diverse LR policies. 
To the best of our knowledge, this is the first to study how to leverage multiple sub-optimal DNNs produced by LR tuning to enhance deep learning predictive performance.

\noindent \textbf{Deep Ensembles}. 
Ensemble learning is a popular Machine Learning (ML) method to combine multiple diverse ML models to improve the overall predictive performance~\cite{dietterich2000ensemble,caruana2004ensemble,bagging,boosting,randomforest,a-survey-ens-learn,wu2021boostingdeep}, represented by bagging~\cite{bagging}, boosting~\cite{boosting}, and random forests~\cite{randomforest}. Deep neural network Ensembles (Deep Ensembles) hold the potential to enhance deep learning predictive performance by combining multiple diverse DNNs~\cite{lakshminarayanan2017simple,deepensembles,wu2021boostingdeep}, where the high diversity is the key to the improved performance. A few existing studies have been devoted to promoting diversity to produce high-quality deep ensembles such as using different random initialization~\cite{lakshminarayanan2017simple} or using different levels of weight decay and label smoothing~\cite{wenzel2020hyperparameter}.
To the best of our knowledge, we are the first to examine whether these models generated during LR tuning can reveal sufficient diversity to facilitate deep ensembles.

\noindent \textbf{Large Language Models}. 
Large Language Models (LLMs) represent a type of auto-regressive generative models trained on massive corpora of texts, which demonstrate remarkable performance to adapt to new tasks and perform human-like conversations merely on textual instructions~\cite{gpt3, llama,instructgpt,llama2}. 
Fine-tuning is an efficient and practical method to further enhance a pre-trained LLM performance for a specific learning task~\cite{alpaca, vicuna}. 
Unlike traditional DNNs, research on LLMs is still in the early stages. In this study, we evaluate LREnsemble using LLMs and demonstrate that \textit{LLM ensembles} can effectively enhance LLM predictive capabilities.

\section{Problem Statement} 

We formally describe the research problems in this section. Following~\cite{LRBenchTIST}, the learning rate tuning is a subproblem of hyperparameter tuning. Concretely, given the optimizer $\mathcal{O}$, deep neural network $F$, loss function $L$, training dataset $X_{\text{train}}$, and validation dataset $X_{\text{validation}}$, the goal of LR tuning is to identify an optimal LR policy $\eta$ to minimize the loss function $L_{x \in X_{\text{val}}}(x; F_{\mathcal{O}_{\eta}(X_{\text{train}})})$, where $F_{\mathcal{O}_{\eta}}$ is the trained DNN by using the optimizer $\mathcal{O}$ and LR policy $\eta$. Formally, the LR tuning problem can be defined as an optimization problem in Formula~\ref{eq:learning_rate_tunning_optimization}:
\begin{equation}
\hat{\eta} = \underset{\eta \in \mathcal{P}}{\mathrm{argmin}} \, L_{x \in X_{\text{val}}}(x; F_{\mathcal{O}_{\eta}(X_{\text{train}})})
\label{eq:learning_rate_tunning_optimization}
\end{equation}
where $\mathcal{P}$ represents a set of all possible LR policies. In practice, LR tuning will identify the optimal LR policy from a finite set of $s$ candidate LR policies, $\mathcal{P} = \{\eta_1(t), \eta_2(t), \ldots, \eta_s(t)\}$. The LR policy $\eta_i(t)$ specifies the LR value for each training iteration/epoch $t$. For example, $\eta_i(t)$ can be a constant value as $\eta_i(t) = 0.1$ for all training iterations/epochs. We can also use a cyclic function, such as a cosine function to define the LR policy, such as $\eta_i(t) = 0.1 \cdot \cos\left(\frac{t}{200} \pi\right)$, a classic cosine learning rate policy with a total of 200 training epochs and 0.1 as the initial LR value. We refer curious readers to~\cite{LRBenchBigData} for the formal definitions of different LR policies.

In order to identify the optimal LR policy, the common practice through trial-and-error is to enumerate all candidate LR policies $\eta_i(t) \in \mathcal{P}$ to train the DNN, producing $s$ model parameters, where each $\theta_i = \mathcal{O}_{\eta_i}$ and $F_{\theta_i}$ is the trained DNN using an LR policy $\eta_i$. The optimal LR policy $\eta^*$ corresponds to the best trained DNN $F_{\mathcal{O}_{\eta^*}}$ evaluated on the validation dataset, which can be easily identified after LR tuning. In addition, we also obtain multiple sub-optimal DNNs trained by other sub-optimal LR policies, i.e., $F_{\mathcal{O}_{\eta_i}}$ where $\eta_i \neq \eta^*$. For simplicity, in total, we obtain $s$ DNNs $F_1$, $F_2$, $\ldots$, $F_s$ from LR tuning including the optimal DNN $F^*$. Often only the optimal LR policy and DNN will be adopted for real-world applications while other sub-optimal DNNs will be discarded, causing a huge waste of computing resources and training time. In this paper, we introduce a new research question: \textit{can we effectively utilize these sub-optimal DNNs to enhance deep learning performance?}

One of the potential solutions is to integrate these diverse models generated during LR tuning, which aligns well with the fundamental principle of deep ensembles~\cite{lakshminarayanan2017simple,deepensembles,wu2021boostingdeep,diversity-and-generalization-in-nn-ensembles}. 
Deep ensembles typically follow three steps to integrate multiple DNNs to potentially enhance the overall predictive performance: (1) base model collection, which creates a base model pool $M$ to include multiple diverse base models through ensemble training or open-source pre-trained models, (2) member model selection, which identifies sub-sets of diverse yet complementary base models to potentially form high-quality deep ensembles, and (3) ensemble consensus, which integrates the outputs of multiple member models to produce the ensemble prediction, such as using soft voting (model averaging) or majority voting. 
Here, the next question is: \textit{will these models trained using different LR policies during LR tuning provide sufficient diversity to serve as ensemble base models?} 
Moreover, given the high variance in the individual model accuracy and a large number of these models from LR tuning (see blue and gray bars in Figure~\ref{fig:first_page_cifar100_lrensemble}), \textit{it is also non-trivial to effectively identify high-quality deep ensembles to boost the overall deep learning predictive performance}.

\section{LREnsemble Overview}  

\begin{wrapfigure}{r}{0.5\textwidth}
\vspace{-10mm}
  \begin{center}
    \includegraphics[width=0.48\textwidth]{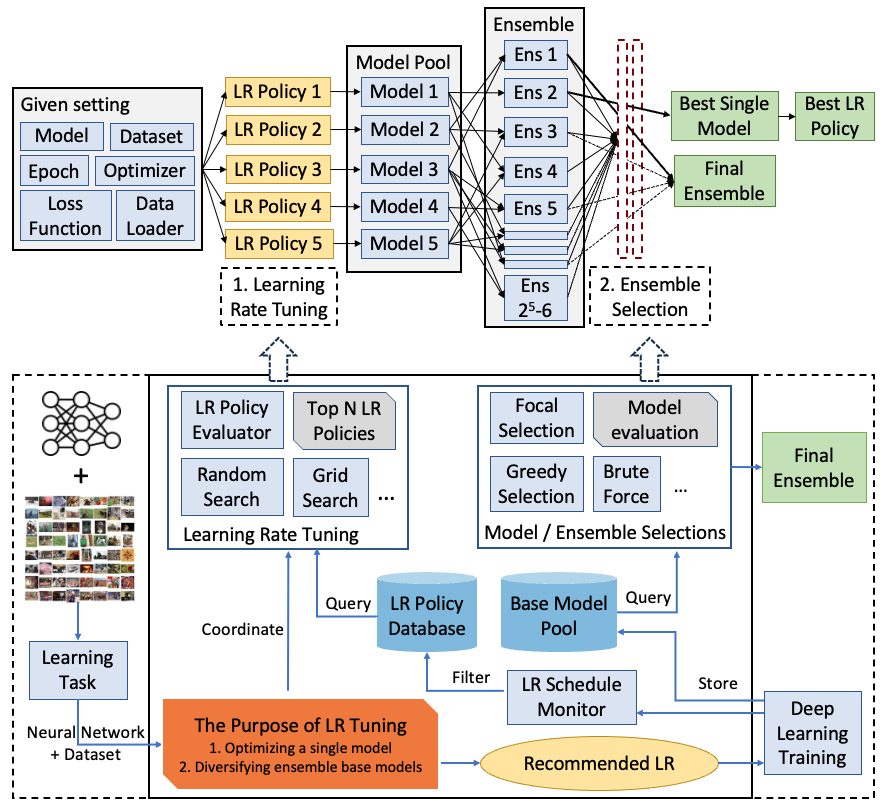}
  \end{center}
  \vspace{-3mm}
  \caption{Overview of LREnsemble architecture}
  \label{fig:lrensemble_framewrok}
    \vspace{-3mm}
\end{wrapfigure}

We propose the LREnsemble framework to address these research challenges by leveraging \textbf{LR} tuning to boost deep \textbf{Ensemble} performance. Figure~\ref{fig:lrensemble_framewrok} presents an overview of LREnsemble. LREnsemble provides functional components to effectively support both LR tuning and deep ensembles. We below describe how to leverage LREnsemble to perform LR tuning and create high-quality deep ensembles.

In LREnsemble, the goals of learning rate tuning include (1) identifying an optimal LR policy to optimize deep learning training and (2) generating diverse individual models to supply the base model pool for building high-quality deep ensembles. These two goals are complementary: by default, we configure LREnsemble with a diverse set of LR policies to explore the large search space of possible LR values, which enables effective identification of the optimal LR policy and meanwhile produces a set of diverse individual models for creating deep ensembles (see our experimental analysis in Section~\ref{section:experimental-analysis}). We implement a variety of popular LR policies, including decaying LRs, cyclic LRs, and composite LRs~\cite{leaderpopulation,autolrs,LRBenchTIST,LRBenchBigData}, and LR tuning strategies, such as grid search, random search, and Bayesian search. In this study, we primarily leverage the grid search and four representative LR policies: (1) MultiStepLR, a popular LR policy defined by a step function~\cite{zagoruyko2016wide,wenzel2020hyperparameter}, (2) WarmupCosineAnnealing (Cosine Annealing LR with Warm-up), a frequently used LR policy, especially for training Transformer-based models~\cite{dosovitskiy2020image, chen2021vision}, (3) OneCycleLR, an LR policy using one cycle of a cyclic function~\cite{onecyclelr-pytorch,onecyclelr-deepspeed} and (4) Composite, a hybrid LR policy consisting of multiple different cyclic functions to facilitate deep learning training at different phrases~\cite{LRBenchTIST}. The LR policy database in LREnsemble stores the LR tuning results organized by the LR policy, dataset, model and learning task, which facilitate the LR tuning to recommend the top $N$ best LR policies.

After LR tuning, multiple diverse individual models will be added to the base model pool, which supplies the base models to build high-quality deep ensembles. Given the base models are generated via LR tuning with varying model accuracy, some of them with low diversity may not complement each other well. Moreover, simply combining all available base models to form the entire ensemble may not produce the optimal accuracy (see Table~\ref{table:ensemble_selection_comparison} in Section~\ref{section:experimental-analysis}). We also provide multiple ensemble selection methods in LREnsemble to effectively identify high quality ensembles with high accuracy, including (1) Brute Force Selection, which enumerates all possible combinations of base models and selects the ensemble with the highest accuracy on the validation set, (2) Greedy Selection, which constructs the ensemble incrementally by initially selecting the best single model, and subsequently adding models that offer the largest performance improvement~\cite{caruana2004ensemble}, (3) Random Selection, which builds the ensemble by randomly choosing base models, potentially subject to certain constraints, such as a fixed team size or a minimum accuracy threshold, and (4) Focal Selection, which leverages focal diversity on the validation set to effectively identify high-quality ensembles~\cite{wu2021boostingensemble}.

LREnsemble is currently implemented with PyTorch~\cite{pytorch}. We adopt a module design for LREns-
emble, which can be easily extended to support new LR policies, new deep learning frameworks, and new ensemble learning methods. LREnsemble can also be integrated with existing general hyperparameter tuning tools, such as Ray Tune~\cite{ray-tune}, to leverage other hyperparameter tuning strategies.

\section{Theoretical Analysis}
In this section, we provide a brief theoretical analysis to explain the improved predictive performance by LREnsemble.
Given SGD~\cite{sgd} as the optimizer for training DNN $F$, we have $\theta_{t+1} = \theta_{t} - \eta_t \nabla L$, where $\theta_{t}$ and $\eta_t$ represent the model parameters and learning rate value respectively at iteration $t$ and $\nabla L$ denotes the gradients. We then analyze the impact of learning rate variance on model parameters. Let $\mu_g$ and $\sigma_g^2$ denote the mean and variance of the gradients $\nabla L$. Assume the learning rate $\eta_t$ has a mean $\mu_{\eta}$ and variance $\sigma_{\eta}^2$. Formally, we have $\mathbb{E}[\nabla L] = \mu_g$, $\text{Var}(\nabla L) = \sigma_g^2$, $\mathbb{E}[\eta_t] = \mu_{\eta}$, and $\text{Var}(\eta_t) = \sigma_{\eta}^2$.



To obtain the variance of model parameters $\text{Var}(\theta_{t+1})$, we use the law of total variance in Formula~\ref{formula:law-of-total-variance}.

\begin{equation}
\text{Var}(\theta_{t+1}) = \mathbb{E}[\text{Var}(\theta_{t+1} \mid \theta_t)] + \text{Var}(\mathbb{E}[\theta_{t+1} \mid \theta_t])
\label{formula:law-of-total-variance}
\end{equation}

Given $\theta_{t+1} = \theta_{t} - \eta_t \nabla L$, we have Formula~\ref{formula:produce-of-independent-variables}.

\begin{equation}
\begin{aligned}
\text{Var}(\theta_{t+1} \mid \theta_t) &= \text{Var}(\theta_{t} - \eta_t \nabla L \mid \theta_t) =  \text{Var}(\eta_t \nabla L) \\
&= \mathbb{E}[\eta_t]^2\text{Var}(\nabla L) + \mathbb{E}[\nabla L]^2\text{Var}(\eta_t) + \text{Var}(\eta_t)\text{Var}(\nabla L) \\
&= \mu_{\eta}^2\sigma_{g}^2 + \mu_{g}^2\sigma_{\eta}^2 + \sigma_{\eta}^2 \sigma_{g}^2
\end{aligned}
\label{formula:produce-of-independent-variables}
\end{equation}

Given $\text{Var}(\eta_t \nabla L)$ does not depend on $\theta_t$, the expectation of the conditional variance is Formula~\ref{formula:expectation-conditional-variance}.

\begin{equation}
\mathbb{E}[\text{Var}(\theta_{t+1} \mid \theta_t)] = \mu_\eta^2 \sigma_g^2 + \mu_g^2 \sigma_\eta^2 + \sigma_\eta^2 \sigma_g^2
\label{formula:expectation-conditional-variance}
\end{equation}

Next, we calculate $\text{Var}(\mathbb{E}[\theta_{t+1} \mid \theta_t])$. Since $\eta_t$ and $\nabla L$ are independent, based on the SGD update rule, we have Formula~\ref{formula:conditional-expectation}.

\begin{equation}
\begin{aligned}
\mathbb{E}[\theta_{t+1} \mid \theta_t] &= \mathbb{E}[\theta_{t} - \eta_t \nabla L \mid \theta_t] = \theta_t - \mathbb{E}[\eta_t \nabla L] \\
&= \theta_t - \mathbb{E}[\eta_t] \mathbb{E}[\nabla L] \\
&= \theta_t - \mu_\eta \mu_g
\end{aligned}
\label{formula:conditional-expectation}
\end{equation}

Hence, the variance of this conditional expectation is Formula~\ref{formula:variance-conditional-expectation}.
\begin{equation}
\text{Var}(\mathbb{E}[\theta_{t+1} \mid \theta_t]) = \text{Var}(\theta_t - \mu_\eta \mu_g) = \text{Var}(\theta_t)
\label{formula:variance-conditional-expectation}
\end{equation}

Combining Formulas~\ref{formula:law-of-total-variance}, \ref{formula:expectation-conditional-variance}, and \ref{formula:variance-conditional-expectation}, we have Formula~\ref{formula:variance-model-parameters}.

\begin{equation}
\text{Var}(\theta_{t+1}) = \mu_\eta^2 \sigma_g^2 + \mu_g^2 \sigma_\eta^2 + \sigma_\eta^2 \sigma_g^2 + \text{Var}(\theta_t)
\label{formula:variance-model-parameters}
\end{equation}

Formula~\ref{formula:variance-model-parameters} shows that the variance of the model parameters $\theta$ depends on both the variance of the learning rate $\sigma_\eta^2$ and the variance of the gradients $\sigma_g^2$, as well as their means. The term $\mu_g^2 \sigma_\eta^2 + \sigma_\eta^2 \sigma_g^2$ highlights that the high variance of the learning rate will contribute to the high variance of model parameters, which can be accumulated across training iterations to significantly increase the variance of model parameters and ultimately produce highly diverse DNNs. 
Moreover, we also utilize diversity-based ensemble selection methods in LREnsemble to further identify highly diverse and complementary DNNs with improved ensemble accuracy and reduced ensemble execution cost.

\section{Experimental Analysis} \label{section:experimental-analysis}
We perform comprehensive experiments to evaluate the proposed LREnsemble. 
The datasets we used in this study include CIFAR-10~\cite{cifar10-100}, CIFAR-100~\cite{cifar10-100}, Tiny ImageNet~\cite{Le2015TinyIV}, and Stanford-alpaca instruction-following data~\cite{alpaca}. The DNN models for evaluation include 
WRN-28-10~\cite{zagoruyko2016wide}, ResNeXt50~\cite{xie2017aggregated}, Vision Transformer (ViT)~\cite{dosovitskiy2020image}, and a large language model, LLaMA-7b~\cite{llama}.
We train the WRN-28-10 model on the CIFAR-10 and CIFAR-100 datasets, and the ResNeXt50 and ViT models on the Tiny ImageNet dataset. We provide the training and dataset details in Section~\ref{appendix:data_setting_details} and~\ref{appendix:training_details} of the appendix. 
We fine-tune LLaMA-7b on the Stanford-alpaca instruct data for three epochs 
with different learning rate policies, and evaluate the fine-tuned models and their ensembles using the metrics of ARC-Challenge~\cite{arc}, Hellaswag~\cite{hellaswag}, and MMLU~\cite{mmlu}.



\subsection{Performance and Diversity of Base Models}

\begin{table*}[ht!]
\vspace{-5mm}
\caption{Accuracy comparison of 16 learning rate policies for training on four tasks}
\label{table:lr-comparison}
\centering
\scalebox{0.65}{
\small
\begin{tabular}{c|ccccccc|cccc}
\hline
\multicolumn{8}{c|}{Learning Rate Policy} & \multicolumn{4}{c}{Accuracy (\%)} \\  \cline{1-12}
\multirow{2}{*}{Learning Rate Function} & \multirow{2}{*}{$k_0$}  & \multirow{2}{*}{$k_0^{vit}$}  & \multirow{2}{*}{$k_1$}   & \multirow{2}{*}{$k_1^{vit}$}  & \multirow{2}{*}{$\gamma$}  & \multirow{2}{*}{$l$} & \multirow{2}{*}{$l_{cycle}$}  &  CIFAR-10 & CIFAR-100 & \footnotesize{Tiny ImageNet} & \footnotesize{Tiny ImageNet} \\ 
& & & & & & & & WRN-28-10 & WRN-28-10 & ResNeXt50 & ViT \\ \hline
MultiStepLR                            & 0.2 & 0.002      &       &   & 0.2     & \footnotesize{[0.3, 0.6, 0.8]}  & & 96.30  & 82.14  & 63.19 & 90.83 \\
MultiStepLR                            & 0.1 & 0.001      &       &   & 0.2     & \footnotesize{[0.3, 0.6, 0.8]} & & 97.12  & 82.66  & 69.31 & \textbf{91.16} \\
MultiStepLR                            & 0.05 & 0.0005    &       &   & 0.2     & \footnotesize{[0.3, 0.6, 0.8]} & & 97.13  & 82.60  & 71.15 & 91.05 \\
MultiStepLR                            & 0.01 & 0.0001    &       &   & 0.2     & \footnotesize{[0.3, 0.6, 0.8]} & & 96.73  & 81.04  & 70.22 & 87.81 \\
OneCycleLR                             & 0.4 & 0.004      & 0     & 0 &         &                                & & 95.86  & 80.82  & 63.36 & 90.21 \\
OneCycleLR                             & 0.2 & 0.002      & 0     & 0 &         &                                & & 96.92  & 83.20  & 68.01 & 90.65 \\
OneCycleLR                             & 0.1 & 0.001      & 0     & 0 &         &                                & & 97.00  & 83.18  & 71.00 & 91.03 \\
OneCycleLR                             & 0.05 & 0.0005    & 0     & 0 &         &                                & & 96.97  & 82.70  & 71.90   & 91.08 \\
WarmupCosineAnnealing                  & 0.2 & 0.002      & 0     & 0 &         &                                & & 96.63  & 83.01  & 66.18  & 90.90  \\
WarmupCosineAnnealing                  & 0.1 & 0.001      & 0     & 0 &         &                                & & 97.04  & \textbf{83.58} & 69.99  & 91.01 \\  
WarmupCosineAnnealing                  & 0.05 & 0.005     & 0     & 0 &         &                                & & \textbf{97.34}  & 83.11  & \textbf{72.41}  & 91.09 \\
WarmupCosineAnnealing                  & 0.01 & 0.0001    & 0     & 0 &         &                                & & 96.82  & 81.38  & 70.49  & 88.83 \\
Composite  & 1.0  & 0.01      & 0.2   & 0.002  & 0.1   & \footnotesize{[0.45, 0.9]} & \footnotesize{[3,2,1]} & 95.40  & 77.85  & 51.07 & 90.28 \\
Composite  & 0.5  & 0.005     & 0.1   & 0.001  & 0.1   & \footnotesize{[0.45, 0.9]} & \footnotesize{[3,2,1]} & 96.31  & 81.94  & 65.71 & 90.75 \\
Composite & 0.25 & 0.0025    & 0.05  & 0.0005 & 0.1   & \footnotesize{[0.45, 0.9]} & \footnotesize{[3,2,1]} & 96.70  & 82.51  & 67.83 & 90.97 \\
Composite & 0.05 & 0.0005    & 0.01  & 0.0001 & 0.1   & \footnotesize{[0.45, 0.9]} & \footnotesize{[3,2,1]} & 97.01  & 81.93  & 71.37 & 90.59 \\ 
\hline
\end{tabular}
} 
\smallskip \\
\raggedright
\footnotesize{The table shows the specific settings we deploy in the training or fine-tuning for each task with different LR policies, including 4 types of LR functions with different settings. $k_0$ is the initial LR value for MultiStepLR and WarmupCosineAnnealing, and is the maximum learning rate for OneCycleLR and Composite LRs. $k_1$ is the minimum LR value. The LR value of the ViT fine-tuning task is shown as $k_0^{vit}$ (=$0.01 \times k_0$) and $k_1^{vit}$ (=$0.01 \times k_1$). The $l$ represents the time/stages when the learning rate multiplies a factor $\gamma$ in terms of the number of iterations/epochs, and $l_{cycle}$ is the cycle length of cyclic LRs for the Composite LRs for the three stages of $l$. All the learning rate policies are visualized in Section~\ref{appendix:lr_policies_title} in the appendix.}
\end{table*}

The first set of experiments investigates whether training models with different learning rate policies in LR tuning can produce diverse individual DNN models.
Table~\ref{table:lr-comparison} presents the experimental results of training DNNs with different LR policies. We highlight three interesting observations. 
\textit{First,} different learning rate policies can generate different training results for all the tasks. Especially, the ResNeXt50 models trained on Tiny ImageNet have up to 21.34\% difference in accuracy. The effect of the learning rate policy is non-negligible and brings diversity to the trained DNN models, which provides potential opportunities for forming high-quality ensembles.
\textit{Second,} the best learning rate policies on three training tasks (WRN-28-10 and ResNeXt50) are WarmupCosineAnnealing policies, while the MultiStepLR policies achieve second and third places in these training tasks and achieve the best in the ViT fine-tuning task. To find the optimal learning rate policy, besides tuning the initial learning rate value, it is worth exploring multiple learning rate policies in depth. 
\textit{Third,} the MultiStepLR policies with an initial learning rate value of 0.1 have a similar performance to another MultiStepLR policy with an initial LR of 0.05, which is not a small gap in the initial LR values. A total of 6 LR policies can train the models to reach 97\% accuracy on the test dataset for CIFAR-10, which shows that different LR policies can train DNNs with similar high accuracy. Hence, LR tuning can potentially produce multiple good LR policies.




\begin{figure}[h!]
\centering
\begin{minipage}{\linewidth}
  \centering
  \subfloat[\scriptsize CIFAR-10 \& WRN-28-10]{
    \includegraphics[width=0.235\linewidth]{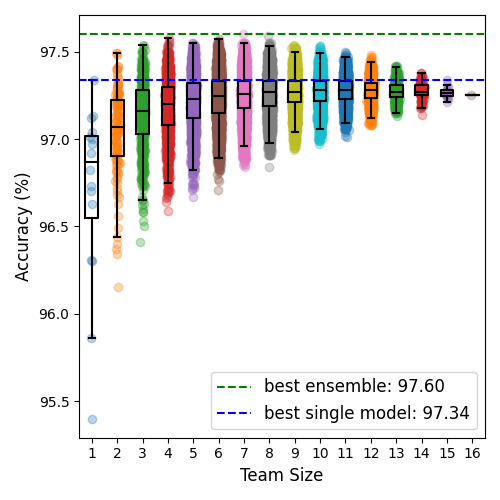}
    \label{fig:cifar10_wrn}
  } 
  \hfill
  \subfloat[\scriptsize CIFAR-100 \& WRN-28-10]{
    \includegraphics[width=0.235\linewidth]{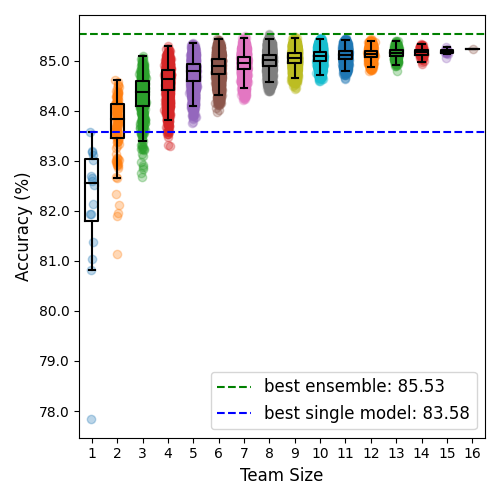}
    \label{fig:cifar100_wrn}
  } 
  \hfill
  \subfloat[\scriptsize Tiny ImageNet \& ResNeXt50]{
    \includegraphics[width=0.235\linewidth]{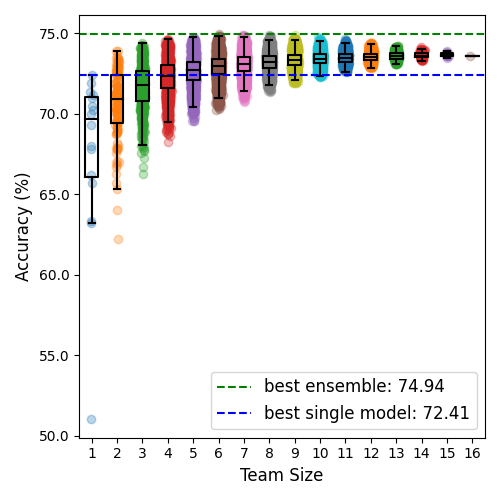}
    \label{fig:tinyimgnet_resnext50}
  } 
  \hfill
  \subfloat[\scriptsize Tiny ImageNet \& ViT]{
    \includegraphics[width=0.235\linewidth]{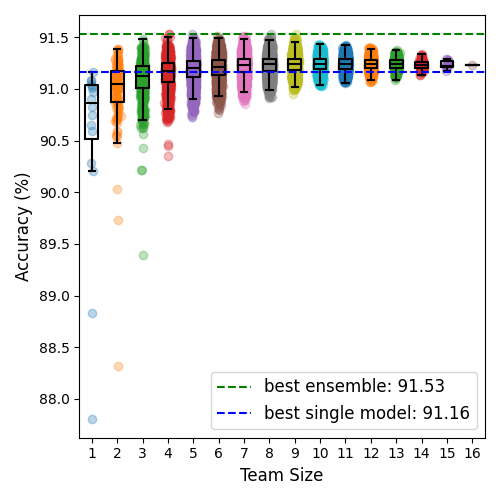}
    \label{fig:tinyimgnet_vit}
  } 
\end{minipage}
\vspace{-2mm}
\caption{Performance distribution across all ensemble team sizes in different tasks.}
\label{fig:ensemble-performance-distribution}
\end{figure}

From previous experiments, we learned that models trained with different LR policies can exhibit varying predictive performance. However, these models produced by LR tuning also have high variance in their accuracy and may not complement each other to boost ensemble performance. 
We then study the potential of leveraging these different models produced by different LR policies to form high-quality ensembles.
Figure~\ref{fig:ensemble-performance-distribution} visualizes the accuracy distributions of every possible ensemble combination for different tasks, where each box presents a performance distribution of a specific team size. A team size of one indicates a single model, while a team size of 16 represents the entire ensemble of all 16 models. We find three interesting observations.
{\it First,} it shows that many ensembles can outperform the best single model for each task, demonstrating the effectiveness of our LREnsemble in creating high-quality ensembles. For training WRN-28-10 on CIFAR-100, the ensembles can consistently outperform the best single model when the team size is larger than 3, a similar case also happens for training ResNeXt50 on Tiny ImageNet when the size is larger than 11.
{\it Second,} among these ensembles, the maximum accuracy shows an increase of 3.4\% in training ResNeXt50 on the Tiny ImageNet dataset and that is 2.33\% in training WRN-28-10 on the CIFAR-100 dataset, which demonstrates the high potential of leveraging LR tuning to effectively boost ensemble performance.
{\it Third,} for all four tasks, we can identify at least one ensemble that consistently outperforms the best single model, which shows that it is beneficial to leverage even sub-optimal models produced in LR tuning to further enhance deep learning performance.
However, the wide distribution of deep ensemble accuracy also presents a critical challenge: many ensembles may not outperform the best single model. Moreover, simply choosing the entire ensemble may not outperform the best single model as well, such as the entire ensemble of 16 WRN-28-10 models on CIFAR-10. Therefore, to fully harness the potential of deep ensembles, we need to identify an effective ensemble selection method, which can efficiently identify complementary ensemble member models to form high-quality ensembles and deliver robust and accurate predictive performance.



\subsection{Performance of Ensemble Selection}
\begin{table}[h!]
\caption{Accuracy comparison of 3 selection methods performance with different team sizes} 
\label{table:ensemble_selection_comparison}
\centering
\scalebox{0.86}{
\small
\begin{tabular}{c|cccc|cccc}
\hline
\multirow{3}{*}{Team Size} & \multicolumn{4}{c|}{Accuracy (\%) of CIFAR-100 \& WRN-28-10} & \multicolumn{4}{c}{Accuracy (\%) of Tiny ImageNet \& ResNeXt50} \\  \cline{2-9} 
 & Random  & Brute  & Greedy  & Focal & Random  & Brute  & Greedy  & Focal \\
 & Selection &  Force &  Selection &  Selection & Selection &  Force &  Selection &  Selection \\ \hline
Best Single  & \multicolumn{4}{c|}{83.58} & \multicolumn{4}{c}{72.41} \\ \cline{1-9}
2  & 84.12$\pm$ 0.32 & 84.52 & 84.52 & 84.52 & 71.24$\pm$ 2.08 & 73.48 & 73.48 & \underline{72.88} \\
3   & 84.07$\pm$ 0.50 & 84.56 & 84.91 & \underline{84.48} & 70.75$\pm$ 1.65 & 74.05 & 74.28 & 73.66 \\
4   & 84.76$\pm$ 0.19 & 84.69 & 84.96 & 84.96 & 72.22$\pm$ 1.41 & 74.06 & 74.22 & 74.27 \\
5   & 84.69$\pm$ 0.34 & 85.02 & 85.12 & 84.91 & 72.76$\pm$ 0.89 & 74.55 & 74.55 & 73.90 \\
6   & 84.77$\pm$ 0.29 & 85.19 & 85.01 & 85.06 & 73.07$\pm$ 0.48 & \textbf{74.75} & \textbf{74.75} & 73.93 \\
7   & 84.86$\pm$ 0.11 & 85.13 & 84.99 & 84.92 & 72.93$\pm$ 0.56 & 74.45 & 74.45 & 74.32 \\
8   & 84.95$\pm$ 0.12 & \textbf{85.53} & 85.17 & 84.93 & 73.13$\pm$ 0.47 & 74.58 & 74.58 & 73.50 \\
9   & 85.04$\pm$ 0.18 & 85.19 & 85.08 & 84.82 & 73.40$\pm$ 0.39 & 74.64 & 74.28 & 74.13 \\
10   & 85.13$\pm$ 0.12 & 85.11 & 84.99 & 85.19 & 73.50$\pm$ 0.39 & 74.33 & 74.33 & 74.14 \\
11  & 85.12$\pm$ 0.15 & 85.17 & 84.95 & 85.32 & 73.27$\pm$ 0.35 & 74.30 & 74.30 & 73.92 \\
12   & 85.11$\pm$ 0.08 & 85.18 & 85.07 & 85.26 & 73.52$\pm$ 0.29 & 74.05 & 74.05 & 73.76 \\
13   & 85.15$\pm$ 0.11 & 85.16 & 85.05 & 85.18 & 73.60$\pm$ 0.22 & 74.16 & 74.16 & 73.43 \\
14   & 85.13$\pm$ 0.06 & 85.12 & 84.99 & 85.22 & 73.57$\pm$ 0.14 & 73.95 & 73.95 & 73.89 \\
15   & 85.20$\pm$ 0.03 & 85.05 & 85.05 & 85.27 & 73.63$\pm$ 0.11 & 73.84 & 73.84 & 73.91 \\ \cline{1-9}   
Entire Ensemble  & \multicolumn{4}{c|}{85.23} & \multicolumn{4}{c}{73.61} \\
\hline
\end{tabular}
}
\end{table}

Given the large pool of base models obtained from LR tuning, random selection or using all models without selection may not produce a good ensemble predictor, since many ensemble teams underperform the best single model as shown in Figure~\ref{fig:ensemble-performance-distribution}. Therefore, we provide multiple ensemble selection methods in our LREnsemble framework.
Table~\ref{table:ensemble_selection_comparison} presents experimental results by using four different ensemble selection methods to select high-quality ensembles using the validation dataset: (1) Random Selection, (2) Brute Force Selection, (3) Greedy Selection~\cite{caruana2004ensemble}, and (4) Focal Selection~\cite{wu2021boostingensemble}. We highlight four interesting observations.
\textit{First,} as depicted in the subplot for training ResNeXt50 on Tiny ImageNet task in Figure~\ref{fig:ensemble-performance-distribution}, when the team size ranges from 2 to 4, around half of the ensembles exhibit lower accuracy than the best single model, with this proportion nearly holding when the team size is 5. Despite the wide range of accuracy distribution and many ensembles underperforming the best single model, all methods except random selection can still select high-quality ensembles. In contrast, random selection fails to achieve similar results under these conditions. Its average performance is lower than the performance of the best single model when the team size is between 2 and 4. 
Moreover, random selection exhibits a lower average performance along with a non-negligible variance compared to the other three selection methods.
\textit{Second,} many ensembles composed of fewer models outperform the entire ensemble of all base models in both tasks, indicating that the best ensemble is often a subset of the entire ensemble, and thus, the ensemble selection or pruning methods would be an important factor for improving the ensemble efficiency and performance. 
~\textit{Third,} as shown in Table~\ref{table:ensemble_selection_comparison}, three selection methods, brute force, greedy selection, and focal selection, have good performance, with no single method proving to be dominant. Considering the focal selection method based on the diversity score, a more explainable and generalization method than the other two methods, we choose this method in the following experiments, and we provide a detailed explanation in the ablation study in Section~\ref{section:ablation-study} in the appendix. 
~\textit{Fourth,} excluding random selection method, the worst case has a 0.9\% and 0.47\% increase compared to the best single model, underlined in Table~\ref{table:ensemble_selection_comparison}. All the ensembles, regardless of the team size, achieve higher accuracy than the best single model. This demonstrates that the models trained by different LR policies with proper ensemble selection can boost overall performance.


\subsection{LREnsemble Performance Comparison to Other Methods}
Our method achieves good performance by leveraging the model diversity produced by learning rate tuning and effective ensemble selections. There are other ensemble learning methods using different diversifying methods or selection methods. Here, we compare our LREnsemble with other state-of-the-art ensemble learning methods for training WRN-28-10 on the CIFAR-10 and CIFAR-100 datasets with 200 epochs.
Table~\ref{table:ensemble-comparison-cifar10-cifar100} presents the performance comparison of the ensemble of size 4 by our LREnsemble and the best ensemble by using other deep ensemble methods, our LREnsemble can consistently outperform other ensemble methods with 0.8\% accuracy improvement over 
\begin{wraptable}{r}{0.68\textwidth} 
\vspace{-3mm}
\caption{Accuracy comparison of LREnsemble and other state-of-the-art methods for training WRN-28-10}
\label{table:ensemble-comparison-cifar10-cifar100}
\centering
\scalebox{0.8}{
\small
\begin{tabular}{c|cc|cc}
\hline
\multirow{2}{*}{Method} & \multicolumn{2}{c|}{CIFAR-10} & \multicolumn{2}{c}{CIFAR-100}  \\
 & Accuracy & Team size  & Accuracy & Team size    \\ 
\hline
Fast Geometric Ensemble~\cite{garipov2018loss}                                    & 96.64    & 12      & 83.12  & 12      \\ 
Snapshot Ensemble~\cite{garipov2018loss, huang2017snapshot}                       & 96.69    & 12      & 83.03  & 12      \\   
BatchEnsemble~\cite{wen2020batchensemble}                                         & 95.94    & 4       & 80.32  & 4       \\
Rank-1 BNN~\cite{dusenberry2020efficient}                                         & 96.5     & 4       & 82.4   & 4       \\
DeepEns~\cite{lakshminarayanan2017simple, wenzel2020hyperparameter}               & 96.2     & 4       & 82.6   & 4       \\   
Hyper-Deep Ens~\cite{wenzel2020hyperparameter}                                    & 96.5     & 4       & 82.8   & 4       \\ \hline
LREnsemble - Focal (4)                                                             & \underline{97.46}   & 4       & \underline{84.96}  & 4      \\
LREnsemble - Focal (best)                                                          & \textbf{97.49}      & 2       & \textbf{85.32}     & 11      \\
\hline
\end{tabular}
}
\vspace{-1mm}
\end{wraptable}
the best method on CIFAR-10 and 2.2\% accuracy enhancement on CIFAR-100. The promising results indicate that LR tuning can produce a pool of diverse and accurate base models, which can be effectively leveraged to build high quality ensembles through ensemble selection. Moreover, this set of experiments also demonstrates that our LREnsemble can effectively leverage LR tuning to boost deep ensemble performance to further enhance deep learning prediction accuracy. 


\subsection{Performance of LREnsemble on LLM Fine-tuning}


With the rapid rise in popularity and influence of Large Language Models, as well as motivated by the encouraging results of fine-tuning ViT in Table~\ref{table:lr-comparison} and Figure~\ref{fig:ensemble-performance-distribution}, which is also a Transformer-based model, we study the performance impacts of LREnsemble on fine-tuning Large Language Models.
Different from the image classification, we leverage another voting method to combine multiple LLMs, which is provided in Section~\ref{section:LLM-ensemble-voting} in the appendix. Table~\ref{table:lrensemble-in-llm} shows the performance comparison of the LLM ensemble and single LLM on ARC-Challenge~\cite{arc}, Hellaswag~\cite{hellaswag}, and MMLU~\cite{mmlu} benchmarks. Fine-tuning can significantly increase the pre-trained LLaMA performance by 4.12\% on ARC-Challenge, 1.43\% on Hellaswag, and 7.28\% on MMLU. The LR tuning by LREnsemble also outperforms the default learning rate setting in Stanford Alpaca~\cite{alpaca} in ARC-Challenge and 
\begin{wraptable}{r}{0.69\textwidth}
\vspace{-2mm}
\centering
\caption{Evaluation results of LLaMA fine-tuning task}
\label{table:lrensemble-in-llm}
\renewcommand{\arraystretch}{1.1}
\scalebox{0.8}{
\small
\begin{tabular}{c|ccc}
\hline
\multirow{2}{*}{Models} & \multicolumn{3}{c}{LLM Evaluation (\%)} \\ 
& ARC(25) & HellaSwag(10) & MMLU(2)  \\ \hline
Origial LLaMA 1~\cite{llama}  & 50.39 & 77.81 & 34.08 \\ 
Fine-tuned LLaMA by Stanford Alpaca~\cite{alpaca} & 52.02 & 78.03 & 41.36 \\ \hline
Best single fine-tuned by LREnsemble  & 54.51 & 79.24 & 41.36 \\ 
Best ensemble fine-tuned by LREnsemble  & \textbf{54.59} & \textbf{79.28} & \textbf{41.80} \\
\hline
\end{tabular}
}
\end{wraptable}
Hellaswag benchmarks, also indicating the high impact of adjusting the learning rates.
Furthermore, the results indicate that the best ensemble team of fine-tuned LLMs identified by LREnsemble can achieve additional performance gains, with an increase of 0.08\% on ARC-Challenge, 0.04\% on Hellaswag, and 0.44\% on MMLU in addition to the improvements obtained from fine-tuning. 
Considering the few iterations in fine-tuning, this improvement shows the potential of our LREnsemble to further boost ensemble performance in training/fine-tuning Large Language Models.


\begin{figure*}[t!]
    \centering
    \includegraphics[width=1\textwidth]{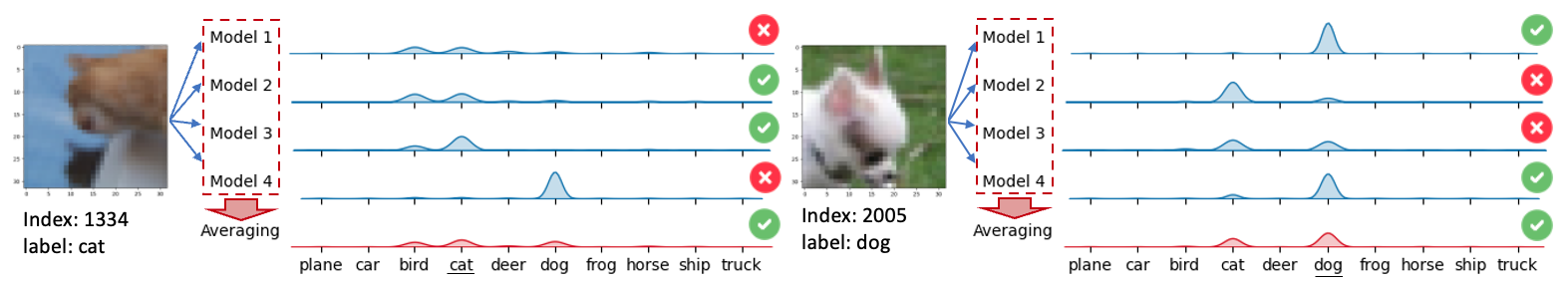}
    \vspace{-5mm}
\caption{Two image examples to illustrate the voting processes in LREnsemble for training WRN-28-10 on CIFAR-10. The ensemble used here is from Table~\ref{table:ensemble-comparison-cifar10-cifar100} with the team size of 4. All four member models are listed in Table~\ref{table:lr-comparison}, which are ordered from model 1 to model 4: MultiStepLR with $k_0$ = 0.1, OneCycleLR with $k_0$ = 0.2, WarmupCosineAnnealing with $k_0$ = 0.1, and WarmupCosineAnnealing with $k_0$ = 0.05.}
\label{fig:lrensemble_demo} %
\vspace{-2mm}
\end{figure*}

\subsection{Visualization of LREnsemble Performance}
Figure~\ref{fig:lrensemble_demo} visualizes the default voting mechanism in LREnsemble. Here, we average the prediction probability vectors from all the member models to compute the final ensemble prediction. Three interesting observations should be highlighted.
\textit{First,} the models trained by using different learning rate policies have significantly different prediction distributions, showing that the different LR policies can introduce diversity to DNNs.
\textit{Second,} this voting mechanism by averaging member model predictions can effectively mitigate biases by individual ensemble members and produce a smoother outcome, which can effectively correct member model mistakes.
\textit{Third,} although two models in the ensemble make incorrect predictions in each case and no single model is correct for both cases, the final ensemble prediction is still correct. This demonstrates that the models within a carefully selected ensemble by LREnsemble can complement each other effectively.


\section{Conclusion}
This paper makes three novel contributions.
\textit{First,} we show that different LR policies can be leveraged to produce highly diverse DNNs to boost deep ensemble performance, which significantly enhances the resource utilization during LR tuning.
\textit{Second,} we propose a novel framework, LREnsemble, to exploit the complementary capability of LR tuning and deep ensembles to improve overall predictive performance.
\textit{Third,} we conduct systematic experiments on multiple benchmark datasets and DNNs, including ViT and LLM models, to demonstrate the effectiveness of LREnsemble. To the best of our knowledge, we are the first to leverage sub-optimal DNNs from LR tuning to improve deep learning accuracy.

\bibliographystyle{ACM-Reference-Format} 
\bibliography{reference}

\newpage
\appendix

\section{Source Codes and Data Setting Details} \label{appendix:data_setting_details}
All the training datasets are split into two components: (1) 90\% for training and (2) 10\% for validation. The data processing contains Random Crop, Random Horizontal Flip, Cutout, and Label Smoothing for all the datasets. In addition, we resize the picture size from 64*64 to 224*224 and add level 2 Random Augment for the Tiny ImageNet dataset.

\section{Visualization of LR Policies} \label{appendix:lr_policies_title}
\begin{figure}[h!]
\centering
\begin{minipage}{\linewidth}
  \centering
  \subfloat[\scriptsize MultiStepLR]{
    \includegraphics[width=0.22\textwidth]{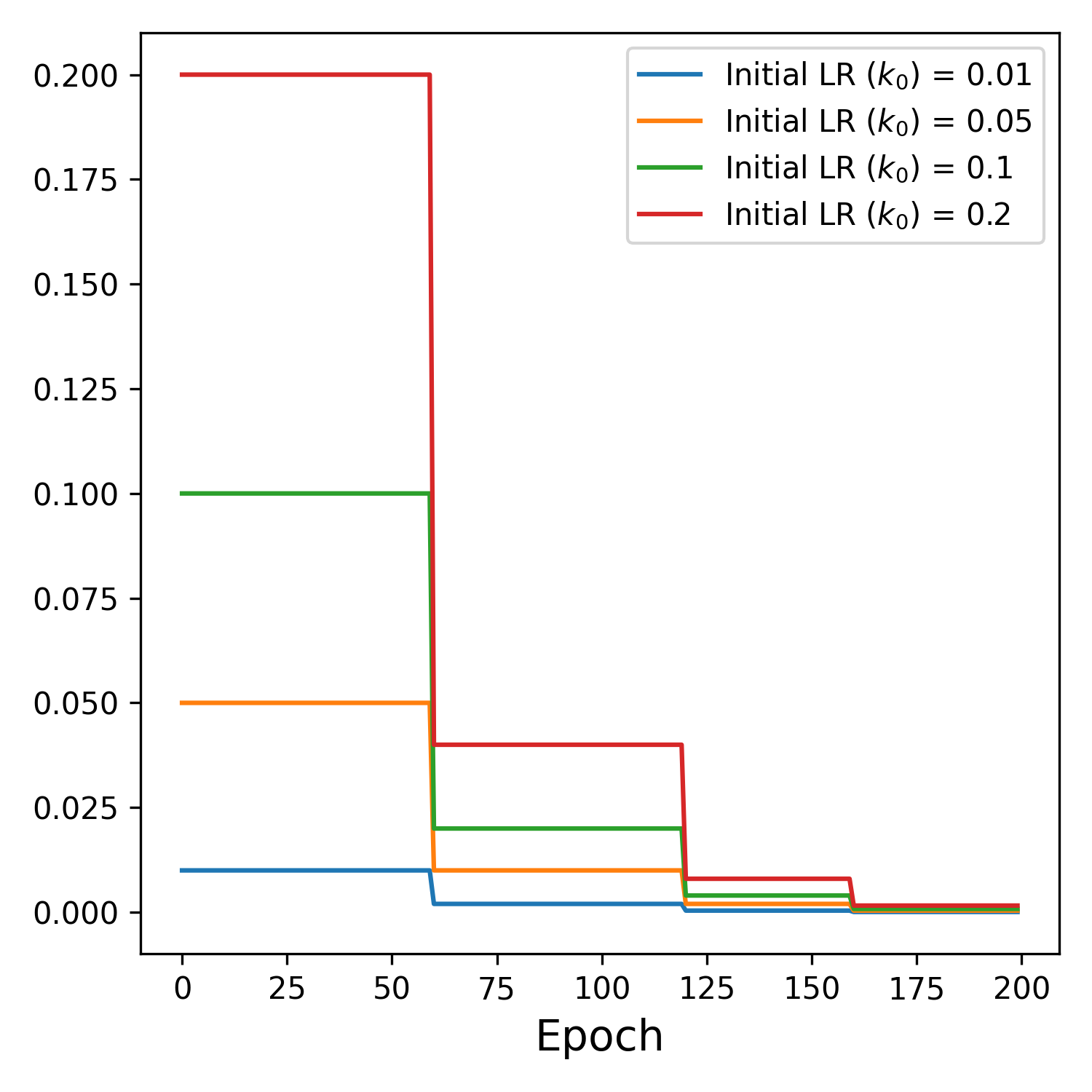}
    \label{fig:piecewise}
  } 
  \hfill
  \subfloat[\scriptsize MultiStepLR (ViT)]{
    \includegraphics[width=0.22\textwidth]{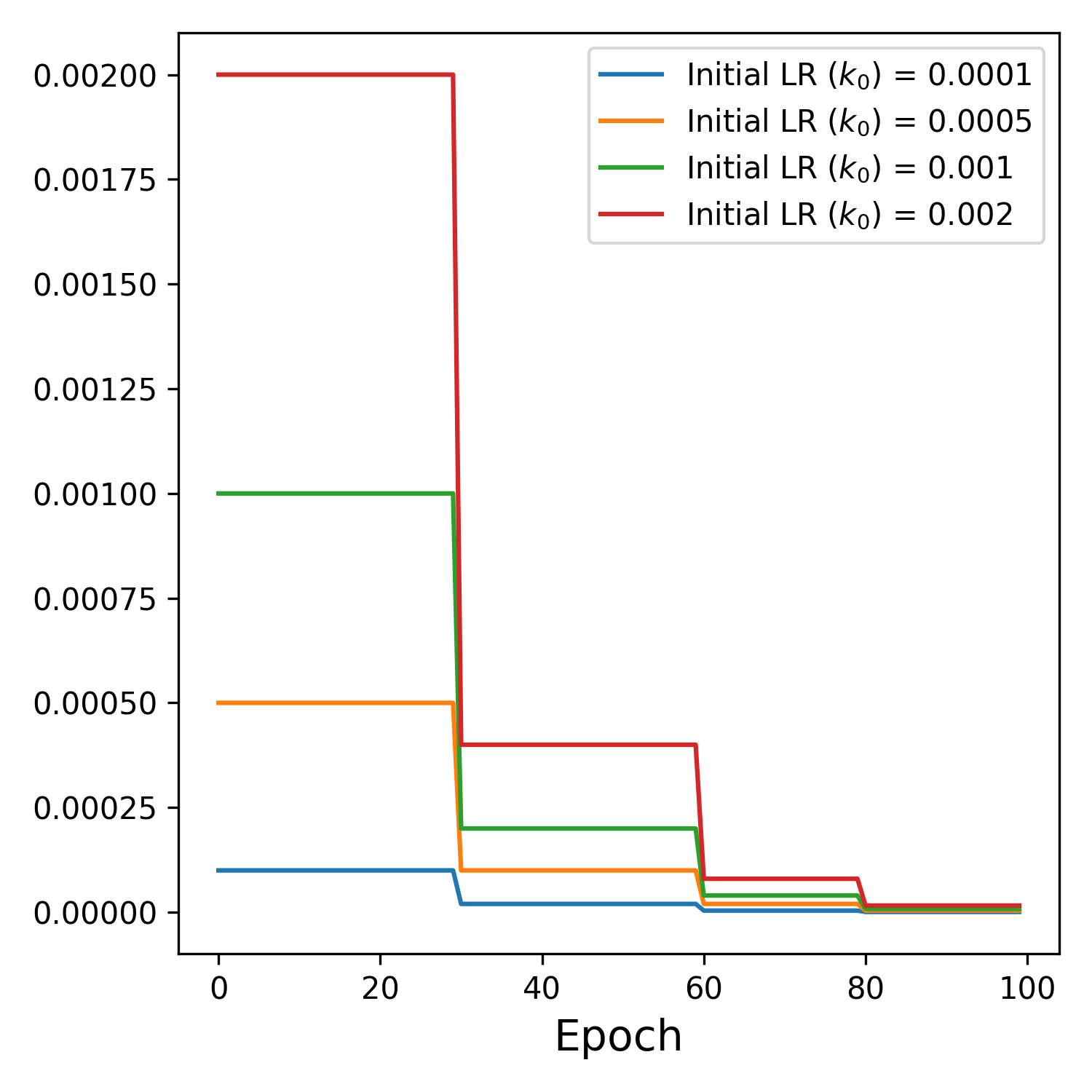}
    \label{fig:piecewise_vit}
  } 
  \hfill
  \subfloat[\scriptsize OneCycleLR]{
    \includegraphics[width=0.22\textwidth]{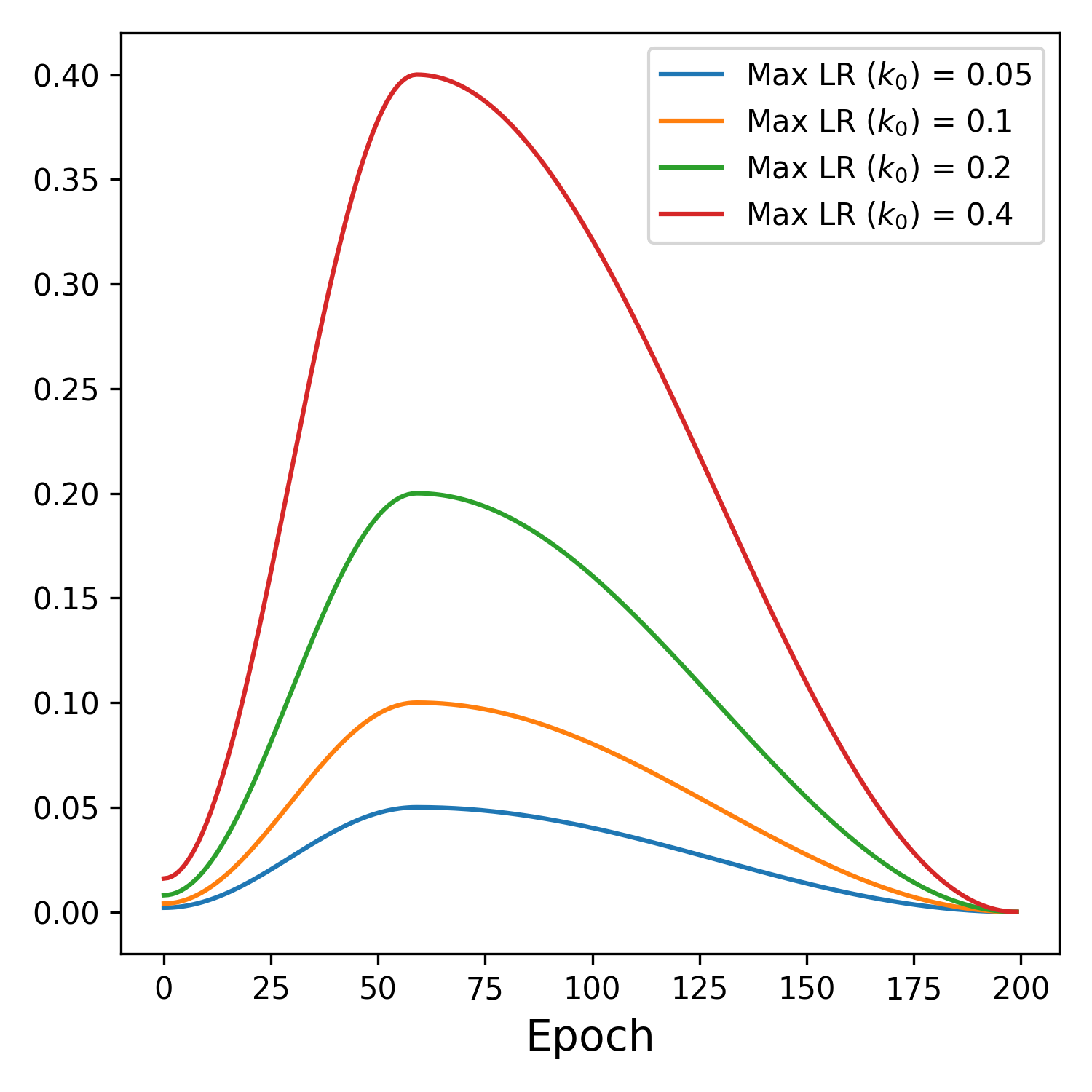}
    \label{fig:onecycle}
  } 
  \hfill
  \subfloat[\scriptsize OneCycleLR (ViT)]{
    \includegraphics[width=0.22\textwidth]{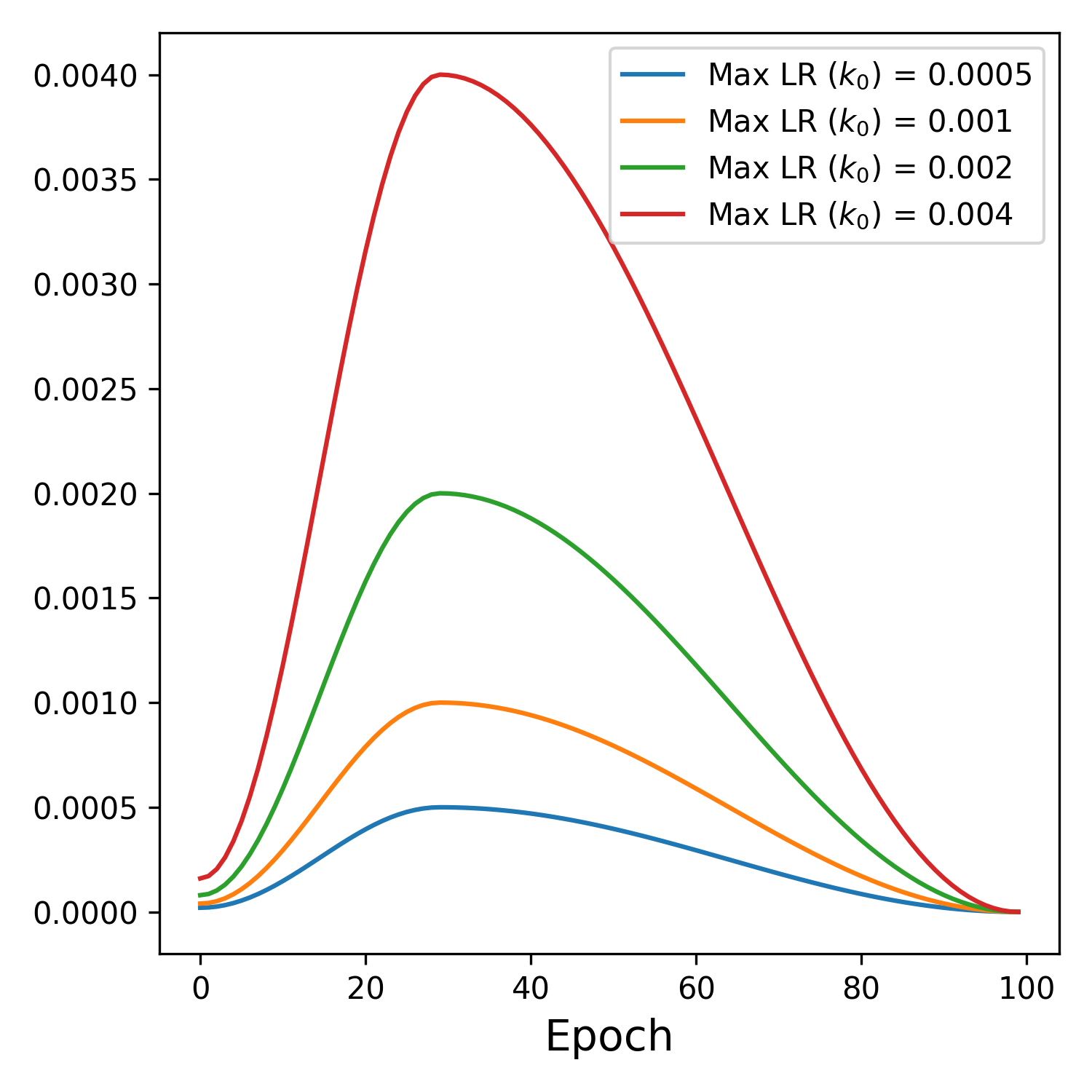}
    \label{fig:onecycle_vit}
  } 
\end{minipage}

\vspace{0.5cm} 

\begin{minipage}{\linewidth}
  \centering
  \subfloat[\scriptsize WarmupCosineAnnealing]{
    \includegraphics[width=0.22\textwidth]{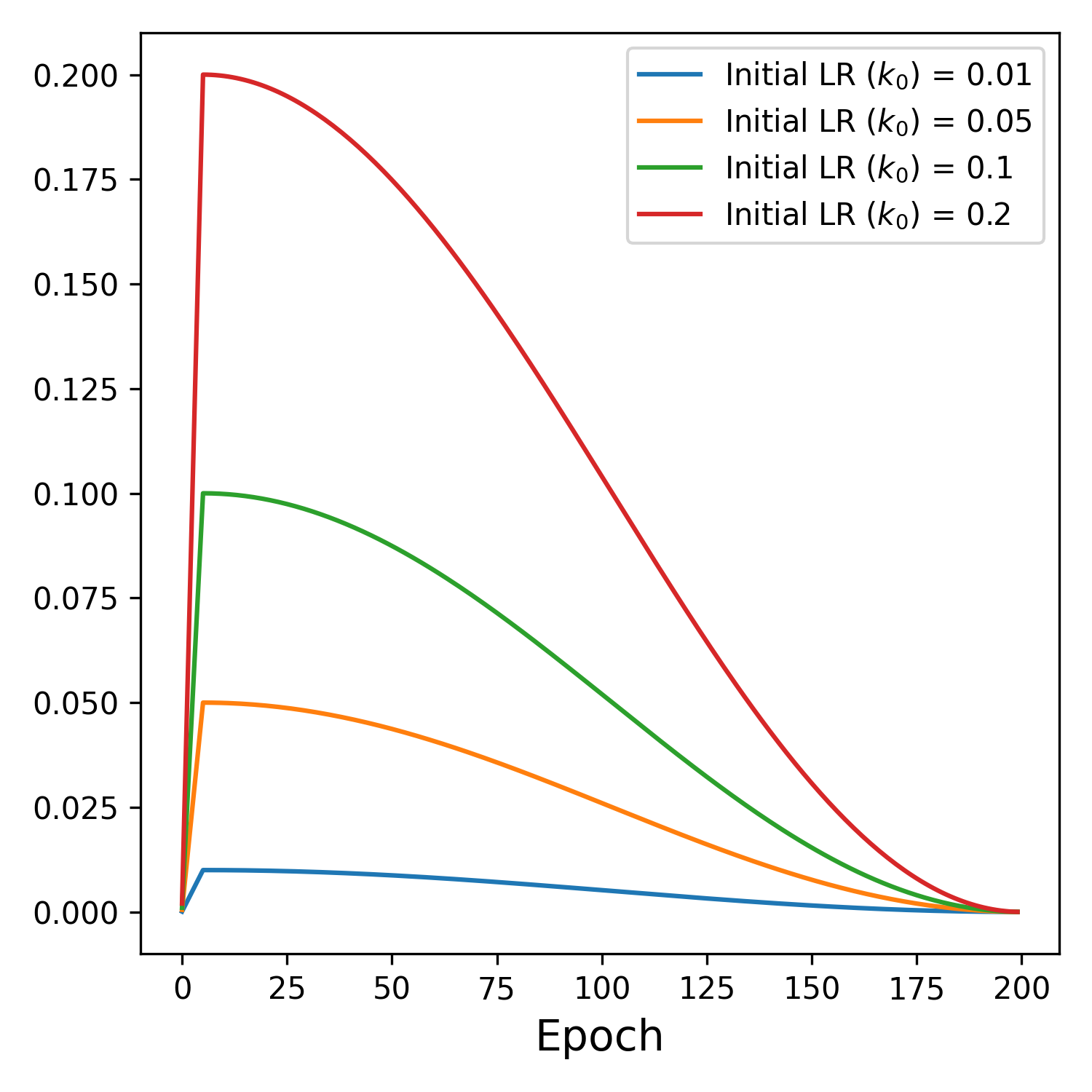}
    \label{fig:cosine}
  } 
  \hfill
  \subfloat[\scriptsize WarmupCosineAnnealing   (ViT)]{
    \includegraphics[width=0.22\textwidth]{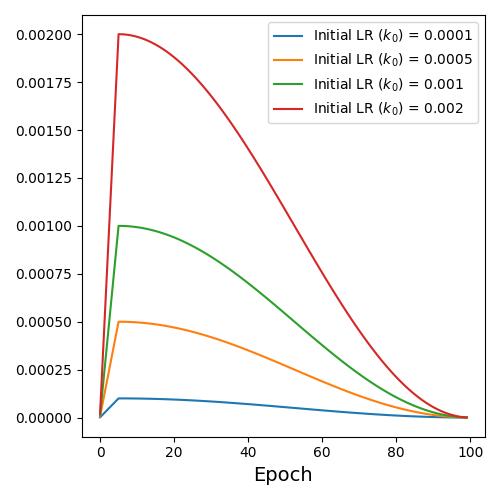}
    \label{fig:cosine_vit}
  } 
  \hfill
  \subfloat[\scriptsize Composite]{
    \includegraphics[width=0.22\textwidth]{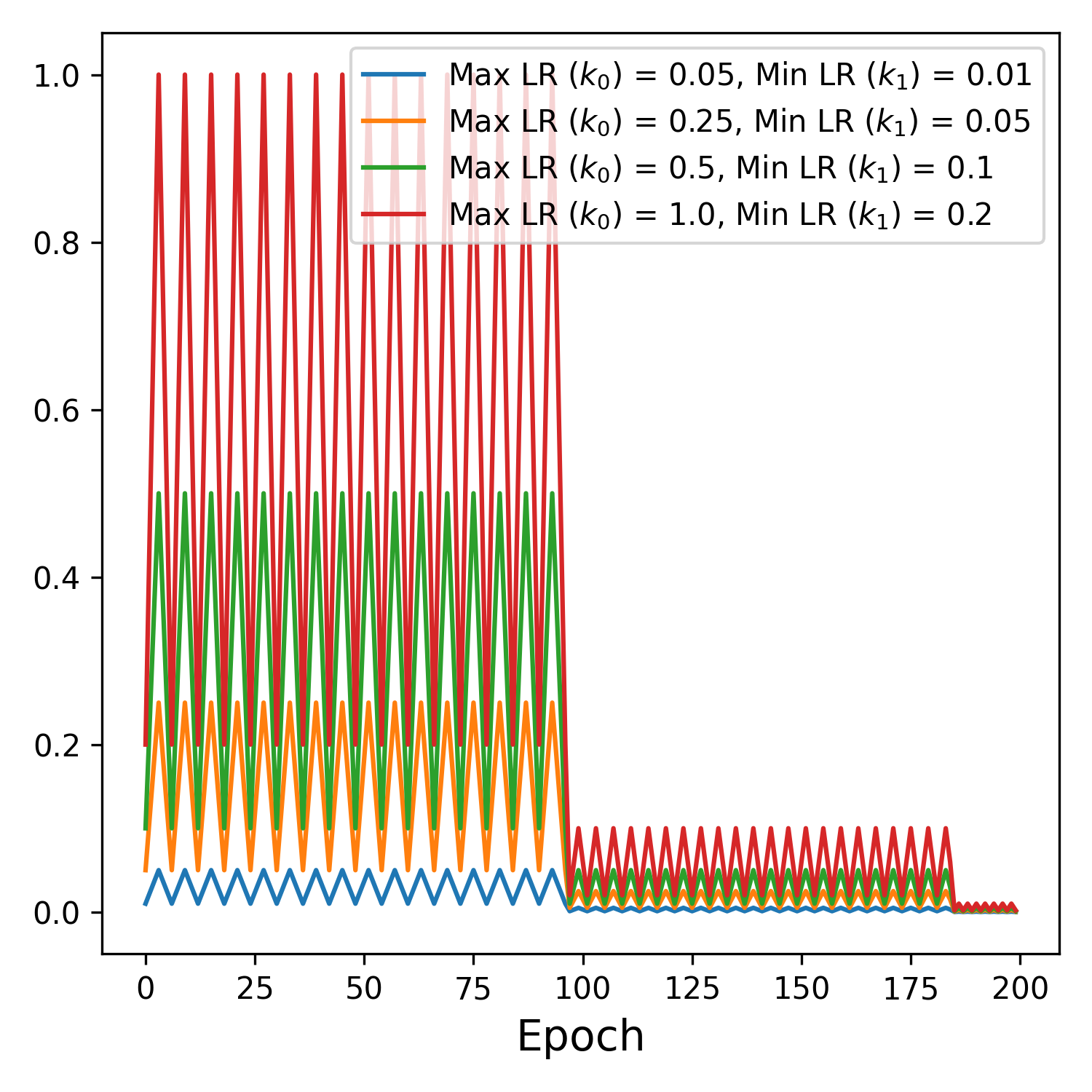}
    \label{fig:composite}
  } 
  \hfill
  \subfloat[\scriptsize Composite (ViT)]{
    \includegraphics[width=0.22\textwidth]{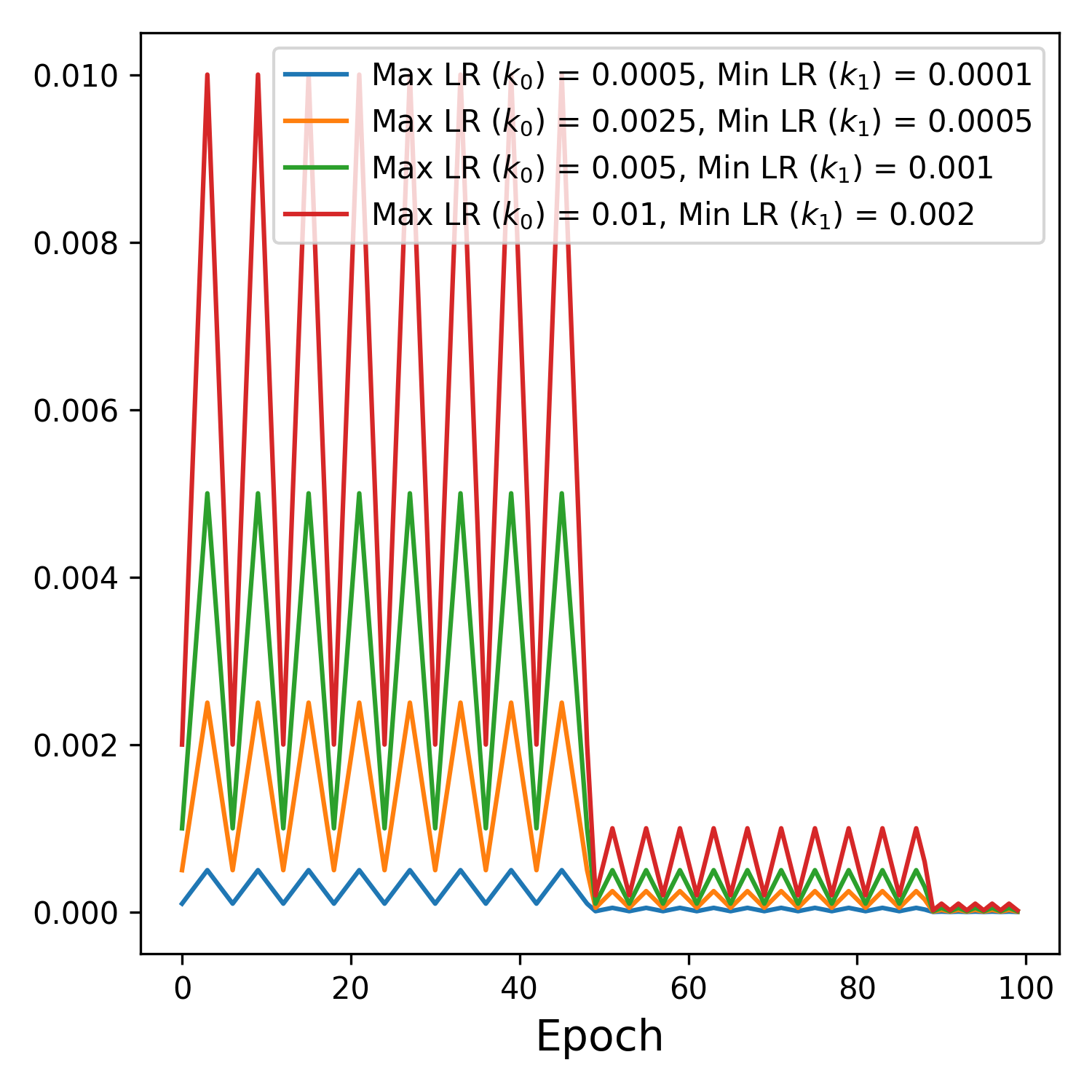}
    \label{fig:composite_vit}
  } 
\end{minipage}

\caption{Visualization of LR policies presented in Table~\ref{table:lr-comparison}.}
\label{fig:lr_policy_visualization}
\end{figure}

\section{Training Details} \label{appendix:training_details}
Most tasks are trained from scratch with the batch size of 64 and 200 total epochs, except ViT. The ViT model is ViT-B/16 pre-trained on ImageNet-1k~\cite{vitb16}, and we fine-tune it on the Tiny ImageNet for 100 epochs with the 256 batch size. All the tasks use ASAM~\cite{foret2020sharpness,kwon2021asam} as the optimizer, setting label smoothing as 0.1, we set all the related seeds as 0 to control the randomness. 

\section{Computational Cost Details} \label{appendix:cost_details}




\begin{table}[h!]
\centering
\caption{Cost comparison for LREnsemble}
\label{tab:cost-analysis}
\scalebox{0.65}{
\begin{tabular}{c|c|c|ccc|ccc|ccc}
\hline
\multirow{3}{*}{Task} &  &  & \multicolumn{9}{c}{Inferencing (sec / million / \%)} \\ 
\cline{4-12}
 & Training & Selection & \multicolumn{3}{c|}{Best Single} & \multicolumn{3}{c|}{LREnsemble} & \multicolumn{3}{c}{Entire Ensemble}\\
 & (min)  & (sec) & Time & \#Params & ACC & Time & \#Params & ACC & Time & \#Params & ACC  \\
 
\hline
CIFAR-10 \& WRN-28-10      & 5279.04    & 2211.86      & 2.26  & 46 & 97.34 &  4.30  & 184 & 97.49 & 34.28  & 736 & 97.25 \\
CIFAR-100 \& WRN-28-10      & 5543.52    & 2965.35      & 2.21  & 46 & 83.58 & 23.62 & 506 & 85.32 & 34.19  & 736 & 85.23 \\
Tiny ImageNet \& ResNeXt50 & 17886.40   & 9115.52   & 5.28  & 25 & 72.41 & 29.55 & 175 & 74.32 & 66.48  & 400 & 73.61 \\
Tiny ImageNet \& ViT       & 34863.52   & 5299.86      & 21.87 & 86 & 91.16 & 65.43 & 258 & 91.37 & 345.36 & 1376 & 91.23 \\
\hline
\end{tabular}
}
\end{table} 


We present a summary of the total computational costs for four training tasks in Table~\ref{tab:cost-analysis}. The experiments are conducted on a server equipped with an Intel(R) Xeon(R) Gold 6258R CPU, 1.5TiB RAM, and an NVIDIA A100 GPU. For the CIFAR-10 and CIFAR-100 datasets, the training/validation/test-
ing sets contain 45,000/5,000/10,000 images respectively.
For the Tiny ImageNet dataset, the training/validation/testing sets consist of 90,000/10,000/10,000 images respectively.

LREnsemble involves selecting models from the base model pool formed during  LR tuning. Consequently, the training time cost reflects the cost of training a single model with LR tuning. Additional costs arise from the model selection and inference phases, as shown in Table~\ref{tab:cost-analysis}. 
These additional costs are considerably minor compared to the training expenses. Moreover, these high-quality ensembles identified by LREnsemble significantly enhance the predictive performance of single models and also substantially reduce the ensemble execution cost by the entire ensemble. 



\section{Ensemble Voting of LLMs} \label{section:LLM-ensemble-voting}
We follow three steps to combine the results from different LLMs fine-tuned by different learning rate policies: 
{\it First,} we get the detailed log file of each model after the evaluation by the Language Model Evaluation Harness~\cite{eval-harness}. We only retain the questions from the three evaluations that are single-choice questions with four options.
{\it Second,} we get the negative likelihood of each question and each model from the logs, and then transform the results of each model in the log using byte-length normalization~\cite{gao2021}. 
{\it Third,} considering the evaluations as four-class classification problems, we use ensemble voting to evaluate the ensemble performance.

\subsection{Performance of LREnsemble on ViT Fine-tuning}


Different from training from scratch, fine-tuning tasks usually only have a small dataset, a few 
iterations/epochs for updating the model parameter, and smaller learning steps~\cite{dosovitskiy2020image,chen2021vision}. 
In our ex-
\begin{wraptable}{r}{0.6\textwidth} 
\vspace{-3mm}
\caption{Performance on ViT fine-tuning task}
\label{table:ensemble_selection_vit}
\centering
\scalebox{0.8}{
\small
\begin{tabular}{c|cccc}
\hline
\multirow{3}{*}{Team Size} & \multicolumn{4}{c}{Accuracy (\%)} \\  
& Random  & Brute   & Greedy   & Focal   \\
&   Selection &   Force &   Selection &   Selection  \\ \hline
2  & 91.00$\pm$ 0.24 & 91.25 & 91.25 & 91.25 \\
3  & 91.04$\pm$ 0.17 & 91.10 & 91.14 & \textbf{91.37} \\
4  & 91.14$\pm$ 0.10 & 91.29 & 91.27 & 91.36 \\
5  & 91.15$\pm$ 0.12 & 91.17 & 91.27 & 91.33 \\
6  & 91.25$\pm$ 0.09 & 91.25 & 91.24 & 91.33 \\
7  & 91.22$\pm$ 0.09 & 91.26 & 91.26 & 91.23 \\
8  & 91.24$\pm$ 0.11 & 91.25 & 91.22 & 91.31 \\
9  & 91.21$\pm$ 0.07 & 91.36 & 91.26 & 91.31 \\
10 & 91.22$\pm$ 0.08 & 91.25 & 91.30 & 91.23 \\
11 & 91.25$\pm$ 0.06 & 91.31 & 91.26 & 91.21 \\
12 & 91.23$\pm$ 0.08 & 91.27 & 91.27 & 91.16 \\
13 & 91.23$\pm$ 0.05 & 91.29 & 91.29 & 91.26 \\
14 & 91.25$\pm$ 0.04 & 91.34 & 91.28 & 91.29 \\
15 & 91.23$\pm$ 0.03 & 91.27 & 91.20 & 91.17 \\ \hline
Entire Ensemble & \multicolumn{4}{c}{91.23} \\ 
\hline
\end{tabular}
} 
\vspace{-3mm}
\end{wraptable}
periments, the ViT fine-tuning only takes half the training time of ResNeXt50 on Tiny ImageNet with learning rates that are two orders of magnitude smaller. 
However, in Table~\ref{table:lr-comparison} and Figure~\ref{fig:ensemble-performance-distribution}, the experimental results of the ViT fine-tuning present similar patterns to the results of the ResNeXt50 training task. 
\textit{First,} Table~\ref{table:ensemble_selection_vit} shows the ViT ensembles identified by LREnsemble can outperform the best single ViT. This observation shows that our LREnsemble is effective in fine-tuning tasks. 
\textit{Second,} beyond the distinctions in model structure and training settings, the performance of the ViT fine-tuning is significantly higher than that of the ResNeXt50 training task, but the effectiveness of LREnsemble is reflected in these different situations, demonstrating the generality of this method. 


\section{Ablation Study} \label{section:ablation-study}
We conduct two ablation studies to better understand the function of learning rate tuning in diversifying models and the role of focal ensemble selection.



\begin{figure}[ht!]
\centering
\subfloat[Model parameter comparison between trained WRN-28-10s]{
  \includegraphics[width=0.33\textwidth]{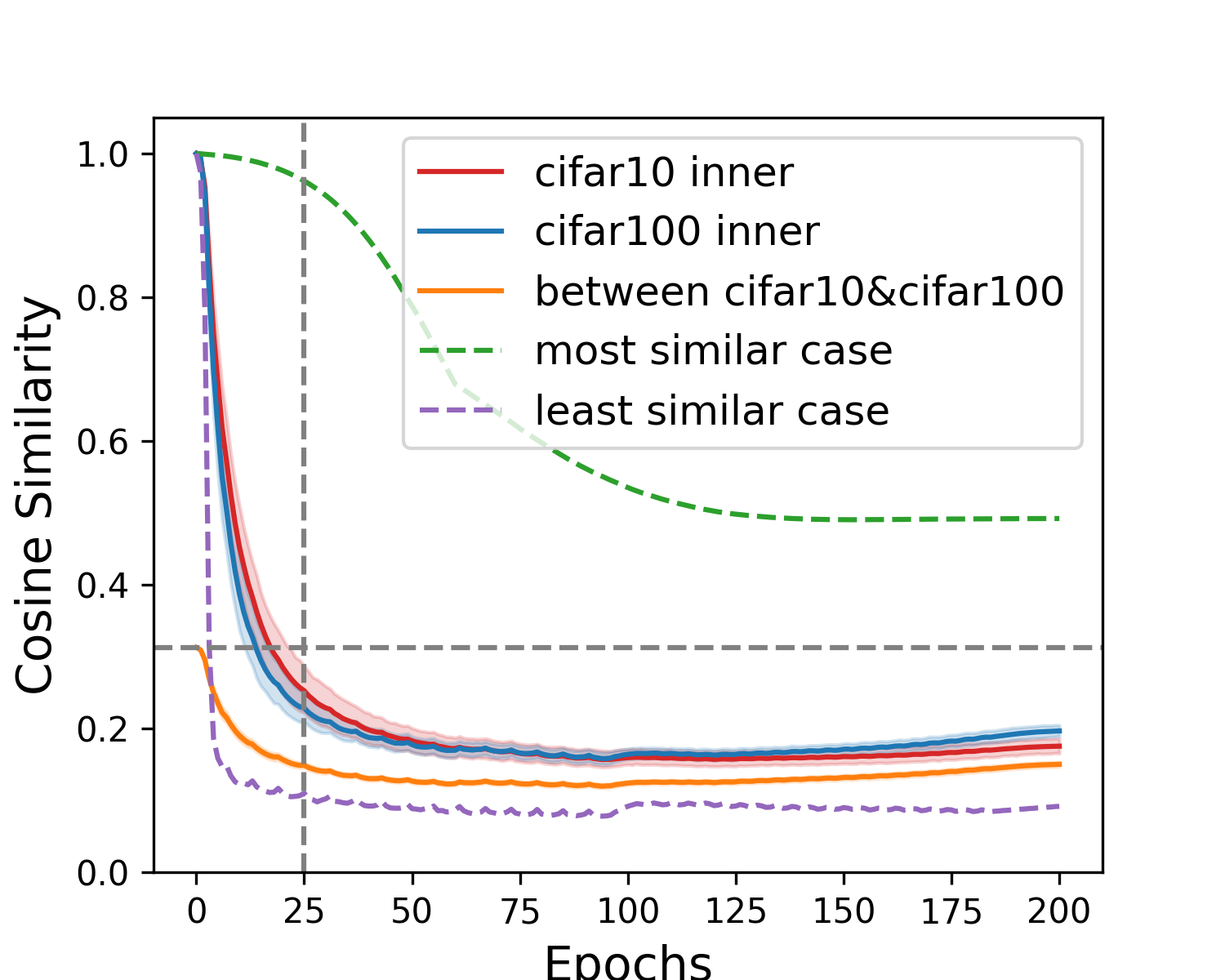}
  \label{fig:model_parameter_comparison}
} 
\subfloat[CIFAR-10 \& WRN-28-10]{
  \includegraphics[width=0.3\textwidth]{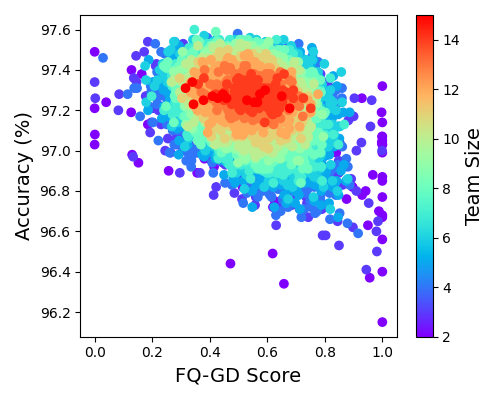}
  \label{fig:cifar10_fq_plot}
} 
\subfloat[Tiny ImageNet \& ResNeXt50]{
  \includegraphics[width=0.3\textwidth]{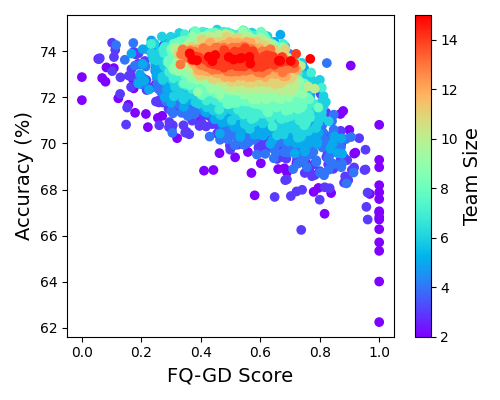}
  \label{fig:resnext50_fq_plot}
} 
\caption{Figure (a) shows the cosine difference between the parameters of each pair of WRN-28-10 models trained on CIFAR-10 or CIFAR-100. Figure (b) and Figure (c) show the relationship between FQ-GD score and accuracy (\%) of the WRN-28-10 Training on CIFAR-10 and the ResNeXt50 Training on Tiny ImageNet.}
\label{fig:combined_figures}
\end{figure}

We calculate the pairwise cosine similarity of the model parameters, except the output layer, for each WRN-28-10 model trained on CIFAR-10 and CIFAR-100, comparing every possible pair of models across all epochs, and summarize the aggregated results shown in Figure~\ref{fig:model_parameter_comparison}. 
The solid lines, accompanied by confidence boundaries, illustrate the trend of cosine similarity scores for model parameters throughout the training period. The red line represents the average value of similarity across all pairs of models trained on the CIFAR-10 dataset, the blue line corresponds to models trained on CIFAR-100, and the orange line depicts the similarity between model pairs with one model trained on CIFAR-10 and the other on CIFAR-100.
The dashed line in green represents the most similar case at the end of the training, and the purple represents the least similar case. Both dashed lines emerged coincidentally from an inner comparison of CIFAR-10.

We find three interesting observations from Figure~\ref{fig:model_parameter_comparison}.
{\it First,} after 25 epochs, the average difference between models with different LR policies is larger than the different model parameter initials in both red and blue lines. Both red and blue lines end at the numbers close to the orange line, which is the pairwise cosine similarity between different initialization and different training tasks. It demonstrates that the cumulative effect of differences in learning rate policies across each iteration can significantly magnify the divergence in model parameters.
{\it Second,} The purple dashed represents the least similar case, which has a 9.16\% similarity score at the end of the training. It is from two WRN-28-10 models trained on CIFRA-10, but the difference exceeds the average difference between WRN-28-10 trained on different datasets, showing the range of model parameter difference brought from learning rate policies can be very large.  
{\it Third,} the most similar cases shown by the green dashed line, from two models trained on CIFAR-10 dataset, only have a 49.21\% similarity score, which shows that the different LR policies will significantly and robustly make the models different and diverse. 
Those above observations, derived from a different analytical perspective compared to the final prediction comparison in Table~\ref{table:lr-comparison}, strongly suggest that the learning rate tuning leads to high diversity in the trained models, underlining the potential for performance enhancement through the ensemble of these models.

We discuss in the experiment part that our LREnsemble method can create a diverse and accurate individual model pool for building a strong ensemble team, which outperforms the best single model in each task and the current methods proposed in other papers in WRN-28-10 model trained on the  CIFAR-10 and CIFAR-100 datasets. By default, we choose the Focal selection in this paper to find the ensemble team for the final evaluation and comparison, even though other selection methods are also applicable based on the result in Table~\ref{table:ensemble_selection_comparison}. 

Both Greedy Selection and Brute Force Selection have generalization problems due to their over-reliance on the validation set~\cite{caruana2004ensemble}, so they might have higher performance on the validation dataset but lack consistent stability and generalizability. However, instead of using the validation set to calculate overall accuracy as the selection metric, the focal selection measures the complementarity among the ensemble members in an ensemble team according to their negative sample independence, in which the FQ-GD score is one of the metrics evaluating the complementarity. Figure~\ref{fig:cifar10_fq_plot} and Figure~\ref{fig:resnext50_fq_plot}  presents a scatter plot that visualizes the relationship between the FQ-GD score and the accuracy for two models: the WRN-28-10 model trained on the CIFAR-10 dataset and the ResNeXt50 model trained on the Tiny ImageNet dataset. There is a negative correlation across all the team sizes between the FQ-GD score and the test accuracy in the graphs in both tasks, showing the good generalization of this method. Based on the plot displayed, we can select the ensemble team with a lower FQ-GD score in a given ensemble team size as our output ensemble.

\end{document}